\documentclass[12pt]{article}

\usepackage[T1]{fontenc}
\usepackage{xcolor}
\usepackage{parskip}
\usepackage{amsmath,amsfonts,amssymb}
\usepackage{centernot,url}
\usepackage[margin=0.75in]{geometry}
\usepackage{bm}
\usepackage[colorlinks=true]{hyperref} 
\usepackage{float}
\usepackage{subfigure}
\usepackage{graphicx}
\usepackage{algorithm}
\usepackage{algpseudocode}
\usepackage{natbib}
\graphicspath{{/home/admin123/Yield_Curve_Modelling/Conference/plots/}}
\usepackage{authblk}

\newcommand{\bfb} {\mbox{\boldmath $\beta$}}

\newcommand{\bfeta} {\mbox{\boldmath $\eta$}}

\newcommand{\bfe} {\mbox{\boldmath $\epsilon$}}

\newcommand{\bft} {\mbox{\boldmath $\theta$}}

\newcommand{\y}{{\mbox{\boldmath $y$}}}

\newcommand{\Y}{{\mbox{\boldmath $Y$}}}

\newcommand{\Z}{{\mbox{\boldmath $Z$}}}

\begin{document}

\title{A Statistical Machine Learning Approach to Yield Curve Forecasting}

\author[1]{Rajiv Sambasivan}
\author[2]{Sourish Das}
\affil[1]{Department of Computer Science, Chennai Mathematical Institute}
\affil[2]{Department of Mathematics, Chennai Mathematical Institute}

\maketitle

\begin{abstract}
\noindent Yield curve forecasting is an important problem in finance. In this work we explore the use of Gaussian Processes in conjunction with a dynamic modeling strategy, much like the Kalman Filter, to model the yield curve. Gaussian Processes have been successfully applied to model functional data in a variety of applications. A Gaussian Process is used to model the yield curve. The hyper-parameters of the Gaussian Process model are updated as the algorithm receives yield curve data. Yield curve data is typically available as a time series with a frequency of one day. We compare existing methods to forecast the yield curve with the proposed method. The results of this study showed that while a competing method (a multivariate time series method) performed well in forecasting the yields at the short term structure region of the yield curve, Gaussian Processes perform well in the medium and long term structure regions of the yield curve. Accuracy in the long term structure region of the yield curve has important practical implications. The Gaussian Process framework yields uncertainty and probability estimates directly in contrast to other competing methods. Analysts are frequently interested in this information. In this study the proposed method has been applied to yield curve forecasting, however it can be applied to model high frequency time series data or data streams in other domains.
\end{abstract}

\section{Introduction and Motivation}
Accurate yield curve forecasting is of critical importance in financial applications. Investors watch the bond market closely as it is a very good predictor of future economic activity and levels of inflation. Future economic activity and levels of inflation affect prices of goods, stocks and real estate. The yield curve is a key representation of the state of the bond market. The slope of the yield curve is an important indicator of short term interest rates and is followed closely by investors. (see \cite{yield_curve_imp}). As a consequence, this has been the focus of considerable research. Several statistical techniques and tools commonly used in econometrics and finance have been applied to model the yield curve (see for example, \cite{Diebold_Li_2006},\cite{Chen_Niu_2014} and \cite{Hays_Shen_Huang_2012}). In this work, we took a machine learning perspective on this problem. Gaussian Processes (GP) are a widely used machine learning technique  (\cite{Rasmussen2005}). We propose a dynamic method that uses Gaussian Processes to model the yield curve.  Yield curve data can be viewed as functional data. Gaussian Process regression has been applied with great success in many domains. Results from this study suggest that Gaussian Process regression performs better than methods currently used for yield curve forecasting in the medium and long term regions of the yield curve. Achieving higher accuracy at longer term structures is more difficult than with the shorter term structures. This is because data points at the longer term structure region of the yield curve are farther apart than in the short term region. A multivariate time series based approach is also commonly used to model the yield curve. This technique had the best results in the short term region of the yield curve. This suggests that these two techniques could be used together. The multivariate time series based approach could be used for short term forecasts and the GP approach could be used for medium and long term forecasting.\\
\noindent The dynamic Gaussian Process method has been applied to model yield curve data in this work. However, functional data presents as a time series in many domains. For example, the hourly user requests processed at a data center could be viewed as functional data. The hourly user traffic for a day may be a variable we wish to forecast. In \cite{das2016understanding}, the daily sea ice surface area in the arctic region observed in one year periods is treated as functional data. Observed sea ice surface area for years passed, could be used to forecast the sea ice surface area for a future year. This suggests that the method proposed in this study could be useful in other application domains too. This is an area of future work.The rest of this paper is organized as follows. In section \ref{sec:fda_review}, we present an overview of relevant aspects of functional data analysis. In section \ref{sec:fcm}, the details of the various methods used to model yield curves, including the proposed method, are provided. In section \ref{sec:test_methodology}, we describe the methodology for validating the performance of the the methods for yield curve forecasting. In section \ref{sec:results}, we describe the results of the study. Finally in section \ref{sec:conclusion}, we present the conclusions from this study. 

\section{Functional Data Analysis, a Review}\label{sec:fda_review}
In functional data, the data have a functional representation.  Yield curve data are represented in terms of the yields associated with a set of term structures. For example, the data for this study consists of 11 terms. We have a yield associated with each term. This constitutes a map (a function) with 11 elements between terms and yield. The  $i^{th}$ yield curve is modeled as a function that maps terms to yields:
$$
y_i=f(\tau_i)+\epsilon_i
$$
The function $f(\tau)$, can be represented as:
\begin{equation}\label{eqn:func_representation}
f(\tau)=\sum_{k=1}^K\beta_k\phi_k(\tau)=\bm{\phi\beta}
\end{equation}
we say $\bm{\phi}$ is a basis system for $f(\tau)$. That is, $$y=\bm{\phi\beta}+\epsilon.$$ Many basis functions have been used for functional representation, each having a particular niche of applications that it is well suited to. 
The sine cosine functions of increasing frequencies
 
\begin{equation}\label{eqn:fourier_basis}
\resizebox{0.5\textwidth}{!}{$
y_i=\beta_1+\beta_2 \sin(\omega \tau) +\beta_3 \cos(\omega \tau) + \beta_4 \sin(2 \omega \tau)+\beta_5 \cos(2 \omega \tau)\hdots +\epsilon_i$}
\end{equation}

forms the Fourier basis, where constant $\omega = 2\pi/P$ defines the period P of oscillation of the first sine/cosine pair. A comparison of Equation \ref{eqn:fourier_basis} with Equation \ref{eqn:func_representation} shows that $$\bm{\phi}=\{1,\sin(\omega \tau),\cos(\omega \tau), \sin(2 \omega \tau), \cos(2 \omega \tau)...\}$$ is the Fourier basis and $\beta^T=\{\beta_1,\beta_2,\beta_3,\hdots\}$ are the corresponding unknown coefficients. Other basis are: 

\begin{itemize}
\item \textbf{Nelson-Siegel Basis}: $\bm{\phi}=\{1,\frac{1-e^{-\lambda\tau}}{\lambda\tau},\frac{1-e^{-\lambda\tau}}{\lambda\tau}-e^{-\lambda\tau}\}$. 
\item \textbf{Exponential Basis}: $\bm{\phi}=\{1,e^{\lambda_1 t},e^{\lambda_2 t}...\}$
\item \textbf{Gaussian Basis}: $\bm{\phi}=\{1,e^{-\lambda (t_1-c)^2},e^{-\lambda (t_2-c)^2}...\}$
\end{itemize}

\subsection{Parameter Learning with Ordinary Least Square Method}
To model functional data, we need to pick a basis function that is appropriate for a particular problem. Once the basis has been picked, the $\beta$'s in Equation \ref{eqn:func_representation} need to be determined. We will discuss the techniques to do this next. One popular technique to determine the $\beta$'s is to use Ordinary Least Squares (OLS). This procedure minimizes the total square error (SSE) between the function and the actual values of the response.
$$
\texttt{SSE}=(\y-\bm{\phi \beta})^T(\y-\bm{\phi \beta}).
$$
The OLS estimate of $\bm{\beta}$ is
$$
\bm{\hat{\beta}}=\bm{(\phi^T\phi)}^{-1}\bm{\phi}^T\y,
$$
and the estimator of $f(t)$ is
$$
\hat{f}=\bm{\phi \hat{\beta}}=\bm{\phi}\bm{(\phi^T\phi)}^{-1}\bm{\phi}^T\y.
$$
The OLS method overfits the model. By overfitting we mean it tries to model the white noise (see \cite{ramsay2002applied} for detailed discussion).

\subsection{Parameter Learning with Penalized Least Square Method}
 
Regularization is one solution to solving the overfitting problem. Regularization achieves this by penalizing the complexity of the solution :
\begin{eqnarray*}
\texttt{PSSE}&=&(y-\bm{\phi \beta})^T(y-\bm{\phi \beta})+\lambda P(f),
\end{eqnarray*}
 $P(f)$ measures the ``roughness" of the $f$, $\lambda$ represents a continuous tuning parameter.
\begin{itemize}
   \item $\lambda \uparrow \infty$ roughness increasingly penalized; $f(t)$ becomes smooth.
   \item $\lambda \downarrow 0$ penalty reduces; $f(t)$ models small shocks and tends to overfit as it move towards OLS.
\end{itemize}
Essentially $P(f)$ measures the curvature of $f(t)$.

\subsection{The $D$ Operator}
We define the $D$-Operator as follows
\begin{itemize}
\item $Df(t)=\frac{\partial}{\partial t}f(t)$ is the instantaneous slope of $f(t)$
\item $D^2f(t)=\frac{\partial^2}{\partial t^2}f(t)$ is the curvature of $f(t)$
\end{itemize}
We measure the size of the curvature for all of $f$ by
\begin{eqnarray*}
P(f)&=&\int [D^2f(t)]^2dt \\ 
&=&\int \bm{\beta}^T [D^2 \bm{\phi}(t)][D^2 \bm{\phi}(t)]^T\bm{\beta}dt\\
&=&  \bm{\beta}^T R_2 \bm{\beta},
\end{eqnarray*}
where $[R_2]_{jk}=\int  [D^2 \phi_j(t)][D^2 \phi_k(t)]^Tdt$ is the penalty matrix. The penalized sum of squares error function is
\begin{eqnarray*}
\texttt{PSSE}&=&(y-\bm{\phi \beta})^T(y-\bm{\phi \beta})+\lambda \bm{\beta}^T R_2 \bm{\beta}
\end{eqnarray*}
Certainly one can try higher order operator as penalty; which is out of the scope of this paper. The penalized least squares estimate for $\bm{\beta}$ is
$$
\hat{\bm{\beta}}=(\bm{\phi^T\phi}+\lambda R_2)^{-1}\bm{\phi}^T\y.
$$
Note that it looks like the `Ridge Estimator'.

\subsection{Parameter Learning with Bayesian Method}
Parameter learning in the methods discussed above involved learning an optimal representation by minimizing a loss function. These approaches posit that that there is a fixed unique set of parameters associated with the functional representation of the yield curve. A contrasting methodology, the Bayesian methodology treats these parameters differently. In Bayesian methodology, the unknown parameters are assumed to be random variables with valid probability measure on the parameter space. 
\subsection{Gaussian Processes} \label{sec:gaussian_process}
Consider the model:
\begin{equation*}
\y = f(t)+\bfe
\end{equation*}
 
Where:
\begin{description}
\item $\bfe \sim \bm{N}(0,\sigma_{\epsilon}^2\bm{I})$. This implies $\y\sim \bm{N}(f(t),\sigma_{\epsilon}^2\bm{I})$.
\item  The function $f(t)$ has the following representation: $$f(t)=\bm{\phi \beta} = \sum_{k=1}^{\infty}\phi_k(t)\beta_k,$$ 
\end{description} 
 
We want to estimate $\bm{\beta}$. We adopt a Bayesian methodology, so we assume $\beta$'s are uncorrelated random variables and $\phi_k(t)$ are known deterministic real-valued functions. Then due to \textbf{Kosambi-Karhunen-Loeve} theorem, $f(t)$ is a \textbf{stochastic process}. 
If we assume $\bm{\beta}\sim \bm{N}(\bm{0},\sigma_{\epsilon}^2\bm{I})$, then $f(t)=\bm{\phi \beta}$ follows a Gaussian process and the induced process on $f(t)$ is known as `\textbf{Gaussian Process Prior}'. The prior on $\bm{\beta}$: $$p(\bm{\beta})\propto  \exp\bigg(-\frac{1}{2\sigma_{\epsilon}^2}\bm{\beta}^T\bm{\beta}\bigg).$$ The induced prior on $f = \bm{\phi \beta}$: $$p(f)\propto  \exp\bigg(-\frac{1}{2\sigma_{\epsilon}^2}\bm{\beta}^T\bm{\phi}^T\bm{K}^{-1}\bm{\phi\beta}\bigg),$$
where the prior mean and covariance of $f$ are given by(see \cite{Rasmussen2005}):
\begin{eqnarray*}
\mathbf{E}[f] &=& \phi E[\bm{\beta}] =\phi \bm{\beta}_0=\bm{0},\\
\mathbf{cov}[f] &=& \mathbf{E}[f.f^T] = \phi. \mathbf{E[\bm{\beta}.\bm{\beta}^T]}\phi^T = \sigma_{\epsilon}^2 \phi .\phi^T = \mathbf{K}.
\end{eqnarray*}
An alternative generic formulation of the model is:
\begin{eqnarray*}\label{eqn:gp_model}
f(\tau)&=&\mu(t)+W(t),\\
\y&=&\mu(t)+W(t)+\bfe,
\end{eqnarray*}
where $W(\tau)\sim \bm{N}(\bm{0},\mathbf{K})$ and $\mu(\tau)$ is a parametric function. 
If there are $m$ many points then,
\begin{eqnarray}
f&\sim& \bm{N}_m\left(\mu(\tau),\mathbf{K}\right), ~~\bfe \sim \bm{N}_m(0,\sigma_{\epsilon}^2\bm{I}_m)\nonumber \\ 
\y&\sim& \bm{N}_m(f(\tau),\mathbf{K}+\sigma_{\epsilon}^2\bm{I}).\label{GP_regression_model}
\end{eqnarray}
The likelihood function is given by:\\
\resizebox{0.4\textwidth}{!}{
$L(f|\y,\bm{\phi},\sigma^2)\propto (\sigma_{\epsilon}^2)^{-m/2}\exp \bigg(-\frac{1}{2\sigma_{\epsilon}^2}(\y-f)^T[\mathbf{K}+\sigma_{\epsilon}^2\bm{I}]^{-1}(\y-f)\bigg),
$}
The negative log-likelihood function can then be expressed as:\\
$$
l(f)\propto \frac{1}{2\sigma_{\epsilon}^2}(\y-f)^T[\mathbf{K}+\sigma_{\epsilon}^2\bm{I}]^{-1}(\y-f).
$$
The corresponding negative log-posterior function is:
$$
p(f)\propto \frac{1}{2\sigma_{\epsilon}^2}\bigg((\y-f)^T[\mathbf{K}+\sigma_{\epsilon}^2\bm{I}]^{-1}(\y-f)+f^T\bm{K}^{-1}f\bigg).
$$
Hence the induced penalty matrix in the Gaussian process prior is identity matrix. It looks like weighted least square method with $L_2$ penalty $P(f)=f^T\bm{K}^{-1}f$. The posterior distribution over these functions is computed by applying Bayes theorem. The posterior is used to make predictions. The estimated value  of $y$ for a given $t$ is the mean (expected) value of the functions sampled from from the posterior at that value of $t$. The expected value of the estimate at $t_*$ is given by:
\begin{align}
\hat{f}(t_*) &= E(f|t_*, \bm{y})\\
 &= \mu(t_*)+ K(t_*, t). [K(t,t) + \sigma_{\epsilon}^2. \bm{I}]^{-1}.(\bm{y}-\mu(t))\label{GP_posterior_mean_for_y}
\end{align}
The variance of the estimate at $t_*$ is given by
\begin{equation}\label{GP_posterior_variance}
cov(f_*) = K(t_*, t_*) - K(t_*, t). [K(t,t) + \sigma_{\epsilon}^2. \bm{I}]^{-1}. K(t, t_*)
\end{equation}

\section{ Forecasting Methods}\label{sec:fcm}
In this section we discuss the methods use to forecast yield curves. This includes the proposed dynamic Gaussian Process method.
\subsection{Nelson-Siegel Model}\label{sec:ns_model}
 
The Nelson-Siegel model \cite{Nelson_Siegel_1987, Chen_Niu_2014} specifies the yield curve as: 
\begin{equation}
\resizebox{0.4\textwidth}{!}{
$y(\tau)=\beta_{1}+\beta_{2}\bigg(\frac{1-e^{-\lambda\tau}}{\lambda\tau}\bigg)+\beta_{3}\bigg(\frac{1-e^{-\lambda\tau}}{\lambda\tau}-e^{-\lambda\tau}\bigg)+\epsilon(\tau), ~~~ \epsilon(\tau)\sim N(0,\sigma_{\epsilon}^2)$}
\end{equation}

where $y(\tau)$ is the yield at maturity $\tau$. The three factors $\beta_{1}$, $\beta_{2}$ and $\beta_{3}$ are denoted as level, slope and curvature of slope respectively. Parameter $\lambda$ controls exponentially decaying rate of the loadings for the slope and curvature. \\
These factors have the following econometric interpretations:
\begin{itemize}
\item The factor $\beta_1$ captures the strength of the long term component of the yield curve.
\item The factor $\beta_2$ captures the strength of the short term component of the yield curve.
\item The factor $\beta_3$ captures the strength of the medium term component of the yield curve.
\end{itemize}
The goodness-of-fit of the yield curve is not very sensitive to the specific choice of $\lambda$ \cite{Nelson_Siegel_1987}. Therefore \cite{Chen_Niu_2014} treated $\lambda$ as a known quantity.
The factors of the Nelson-Siegel model need to be estimated from the data for the yield curve. Yield curve data are instances of a type of data called functional data. When this technique is applied to a successive yield curves, there could be a pattern in the evolution of the coefficients for the Nelson-Siegel model over time. Section \ref{sec:ts_ns} provides a mathematical framework to abstract this problem.

\subsection{Multivariate Time Series Forecasting}\label{sec:ts_mv}
A common method to model yield curve data is to use a Vector Auto-Regressive model to represent the yields for the term structures. (\cite{diebold2013yield}). An auto-regressive model of order $k$ is represented by:
\begin{equation}\label{eqn:ar_model}
\y_i(\tau)  = \beta_0 + \beta_1.\y_{i-1}(\tau) + \hdots + \beta_k\y_{i-k}(\tau)  
\end{equation}
Equation \ref{eqn:ar_model} represents a regression of the $i^{th}$ yield curve on the previous $k$ yield curves. A model selection criterion, like the Bayesian Information Criterion is used to determine the optimal order, $k$,  for the data. Forecasting is then performed using the optimal model. The results of modeling are presented in section \ref{sec:results}. 
\subsection{Forecasting the Yield Curve through Nelson-Siegel Parameters}\label{sec:ts_ns}

The Dynamic Nelson-Siegel (DNS) model \cite{Nelson_Siegel_1987, Chen_Niu_2014} for yield curve has the following representation:\\
\resizebox{0.9\linewidth}{!}{
\begin{minipage}{\linewidth}
\begin{align*}
y_t(\tau_j) &= \beta_{1t}+\beta_{2t}\bigg(\frac{1-e^{-\lambda\tau_j}}{\lambda\tau_j}\bigg)+\beta_{3t}\bigg(\frac{1-e^{-\lambda\tau_j}}{\lambda\tau_j}-
e^{-\lambda\tau_j}\bigg)+\epsilon_t(\tau_j),\\
\beta_{it} &= \theta_{0i}+\theta_{1i}\beta_{i,t-1}+\eta_{i},~~i=1,2,3
\end{align*}
\end{minipage}
}

here:
\begin{itemize}

\item $\epsilon_t(\tau_j)\sim N(0,\sigma_{\epsilon}^2)$
\item$\eta_i\sim N(0,\sigma_{\eta}^2)$,
\item$t=1,2,\hdots, T$ represents the time steps in days
\item $j=1,2,\hdots,m$ represents the term structure or maturity
\item$y_t(\tau)$ is the yield for maturity $\tau$ (in months) at time $t$. 
\end{itemize}
The three factors $\beta_{1t}$, $\beta_{2t}$ and $\beta_{3t}$ are denoted as level, slope and curvature of slope respectively. Parameter $\lambda$ controls exponentially decaying rate of the loadings for the slope and curvature. The goodness-of-fit of the yield curve is not very sensitive to the specific choice of $\lambda$ \cite{Nelson_Siegel_1987}. Therefore \cite{Chen_Niu_2014} chose $\lambda$ to be known. In practice, $\lambda$ can be determined through grid-search method. There are eight static parameters $\bft=(\theta_{01},\theta_{02},\theta_{03},\theta_{11},\theta_{12},\theta_{13},\sigma_{\epsilon}^2,\sigma_{\eta}^2)$ in the model. In matrix notation the DNS model can be presented as 
\begin{eqnarray}
\bfb_{t}&=&\theta_0+\Z\bfb_{t-1}+\bfeta_t, \label{system_equation}\\
\y_t&=&\bm{\phi}\bfb_t+\bfe_t,\label{observation_equation}
\end{eqnarray}
where
$\y_t=\left(\begin{array}{c}
y_t(\tau_1)\\
y_t(\tau_2)\\
\vdots\\
y_t(\tau_m)
\end{array}
\right)_{m \times 1}$,\\
$\bm{\phi}=\left(\begin{array}{ccc}
1& f_1(\tau_1) & f_2(\tau_1)\\
1& f_1(\tau_2) & f_2(\tau_2)\\
\vdots & \vdots & \vdots \\
1& f_1(\tau_m) & f_2(\tau_m)\\
\end{array}
\right)_{m \times 3}$,\\
$
\bfb_t=\left(\begin{array}{c}
\beta_{0t}\\
\beta_{1t}\\
\beta_{2t}
\end{array}
\right)_{3 \times 1},
$
$\bfe_t=\left(\begin{array}{c}
\epsilon_1\\
\epsilon_2\\
\vdots\\
\epsilon_m
\end{array}
\right)_{m \times 1}$,\\

\noindent such that:
\begin{itemize}
\item  $f_1(\tau_j)=\big(\frac{1-e^{-\lambda\tau_j}}{\lambda\tau_j}\big)$
\item $f_2(\tau_j)=\big(\frac{1-e^{-\lambda\tau_j}}{\lambda\tau_j}-e^{-\lambda\tau_j}\big)$, $j = 1,2,...,m$. The index $j$ represents the term structure or maturity. There are $11$ term structures for this study ($m = 11$)
\item $\theta_0=\left(\begin{array}{c}
\theta_{01}\\
\theta_{02}\\
\theta_{03}
\end{array}
\right)$
\item  $\Z=\left(\begin{array}{ccc}
\theta_{11}&0&0\\
0&\theta_{12}&0\\
0&0&\theta_{13}
\end{array}
\right)$
\end{itemize}

Note that $\bfe_t \sim \bm{N}_m(0,\sigma_{\epsilon}^2\bm{I}_m)$ and $\bfeta_t \sim \bm{N}_3(0,\sigma_{\eta}^2\bm{I}_3)$. Note that (\ref{system_equation}) is \textit{system equation} and (\ref{observation_equation}) is \textit{observation equation}. \cite{Diebold_Li_2006} suggest that the factors of the Nelson-Siegel model be estimated using a least squares procedure. The data for each yield curve produces a set factors associated with the Nelson-Siegel representation of the yield curve. The dataset is a collection of yield curves. Therefore sequential application of the least squares procedure would yield a set of Nelson-Siegel factors. The evolution of these factors can be represented using a Vector Auto-Regressive model. A model selection methodology like the Bayesian Information Criterion can be used to determine the optimal lag order for the model. Once an optimal model structure has been determined, forecasting is performed using the optimal model.

\subsection{Forecast with Dynamic Gaussian Process Prior Model} \label{sec:dgp}

Here we introduce dynamic Gaussian process prior model. The observation equation is
\begin{eqnarray*}
\y_t &=&\mu_t(\tau)+\bfe_t,\label{DGP_observation_equation}
\end{eqnarray*}
where $\y_t$ and $\bfe_t$ are defined as in (\ref{observation_equation}), $\mu_t(\tau)$ is the mean function. The system equation is defined as
\begin{eqnarray}
\mu_t(\tau) &=&\mu_{t-1}(\tau)+W_t,\label{GP_system_equation}
\end{eqnarray}
where $W_t(\tau) \sim \bm{N}_m(\bm{0},\mathbf{K}_{t-1})$, where $\mathbf{K}_{t-1}=K(\tau,\tau'|\rho_{t-1})$, $\rho_{t-1}$ is the hyper-parameter estimated at $t-1$. The key notion here is that given the data $\Y_t = (\y_t,\y_{t-1},\hdots,\y_1)$ inference about $\mu_{t}$ and prediction about $\y_{t+1}$ can be carried via Bayes theorem, which can be expressed as
\begin{eqnarray}
\mathbb{P}(\mu_{t}(\tau)|\Y_t)\propto \mathbb{P}(\y_t|\mu_{t}(\tau),\Y_{t-1})\times \mathbb{P}(\mu_{t}(\tau)|\Y_{t-1}).\label{Bayes_theorem_GP}
\end{eqnarray}
Note that the expression on the left of equation (\ref{Bayes_theorem_GP}) is the \textit{posterior process} of $\mu(\tau)$ at time $t$, whereas the first and second expression on the right side of (\ref{Bayes_theorem_GP}) is the \textit{likelihood} and \textit{prior process} of $\mu(\tau)$, respectively. Suppose the posterior process at time point $t-1$ is the 
\begin{eqnarray}
\mu_{t-1}|\Y_{t-1}\sim \bm{N}_m\left(\hat{\mu}_{t-1}(\tau),\hat{\mathbf{K}}_{t-1}\right), \nonumber 
\end{eqnarray}
where $\hat{\mu}_{t-1}(\tau)$ is the posterior mean function and $\hat{\mathbf{K}}_{t-1}$ is the posterior covariance function of the process at the time-point ($t-1$). Following the structure of the GP regression model as presented in (\ref{GP_regression_model}) and (\ref{GP_system_equation}), the prior predictive process at time point $t$ is
\begin{eqnarray}
\mu_t|\Y_{t-1}&\sim& \bm{N}_m\left(\hat{\mu}_{t-1}(\tau),\hat{\mathbf{K}}_{t-1}\right), \nonumber 
\end{eqnarray}
the likelihood function is
\begin{eqnarray}
\y_t |\mu_{t}(\tau),\Y_{t-1}\sim \bm{N}_m(\mu_t(\tau),\sigma_{t}^2\bm{I}_m),\nonumber 
\end{eqnarray}
and the marginal likelihood function is
\begin{eqnarray}
\y_t|\Y_{t-1} &\sim& \bm{N}_m(\hat{\mu}_{t-1}(\tau),\hat{\mathbf{K}}_{t-1}+\sigma_{t-1}^2\bm{I}_m).\label{marginal_likelihood_GP}
\end{eqnarray}
Note that in (\ref{marginal_likelihood_GP}) the $\mu_{t-1}(\tau)$ is a measurable under the $\sigma$-field generated by $\Y_{t-1}$. We can estimate the hyper-parameters $\theta_t=(\rho_{t-1},\sigma_{t-1})$, using a optimization procedure to  maximize the marginal-likelihood (\ref{marginal_likelihood_GP}). Let's assume $\hat{\theta}_{t-1}$ is the estimated hyper-parameter estimated by optimizing the (\ref{marginal_likelihood_GP}). We can then provide an estimate for the observation at time $t$ using the expected value of $\y_t|\Y_{t-1}$ (obtained from \ref{marginal_likelihood_GP}). This is:
\begin{align*}
\hat{\mu}_{t}(\tau*) &=\mathbb{E}(\mu_t(\tau*)|\Y_{t-1})\\
&= K(\tau*, \tau|\hat{\rho}_{t-1}). [K(\tau,\tau|\hat{\rho}_{t-1}) + \hat{\sigma}_{t-1}^2. \bm{I}]^{-1}.\y_{t-1}(\tau).
\end{align*}
Once we have obtained the observation at time $t$, we can update the posterior process over $\y_t$ as:
\begin{eqnarray}
\y_t(\tau)|\Y_t \sim \bm{N}_m(\hat{\mu}_{t.updated}(\tau),\hat{\mathbf{K}}_{t.updated}),\nonumber
\end{eqnarray}
where the corresponding covariance function is
\begin{equation*}
\resizebox{0.4\textwidth}{!}{
$\hat{\mathbf{K}}_{t.updated} = K(\tau_*, \tau_*|\hat{\rho}_{t}) - K(\tau_*, \tau|\hat{\rho}_{t}). [K(\tau,\tau|\hat{\rho}_{t}) + \hat{\sigma}_{t}^2. \bm{I}]^{-1}. K(\tau, \tau_*|\hat{\rho}_{t})$},
\end{equation*}
and the mean function or the expected value of $\mu_t$ at $\tau*$ is:\\
\resizebox{0.9\linewidth}{!}{
\begin{minipage}{\linewidth}
\begin{align*}
\hat{\mu}_{t.updated}(\tau*) &=\mathbb{E}(\hat{\mu}_t(\tau*)|\Y_t)\\
 &=\hat{\mu}_{t}(\tau*)+K(\tau*, \tau|\hat{\rho}_t). [K(\tau,\tau|\hat{\rho}_t) + \hat{\sigma}_{t}^2. \bm{I}]^{-1}.\big(\y_{t+1}-\hat{\mu}_{t}(\tau)\big).
\end{align*}
\end{minipage}
}

The details of an algorithmic implementation of this procedure is provided section \ref{sec:dgp_alg}

 \subsubsection{The Dynamic Gaussian Process Algorithm} \label{sec:dgp_alg}
There are two distinct phases of the algorithm. These correspond to the time steps $t = 0$ (the first yield curve in the dataset) and $ t > 0$ (the subsequent yield curves in the dataset). The details of each of these phases is provided below.

\noindent \textbf{Time step $t=0$}:
\begin{enumerate}
\item \textbf{Hyper-parameter Estimation:} Estimate hyper-parameters, $\hat{\theta}_0$,  of the Gaussian Process $\y_0 \sim \bm{N}_m(\bf{0},\mathbf{K}+\sigma_{0}^2\bm{I}_m)$. Here $\mathbf{K}= K(\tau,\tau*|\rho_0)$ and $\theta_0=(\rho_0,\sigma_{0})$ are the hyper-parameters at time $t = 0$. The hyper-parameters are obtained by maximizing the marginal log-likelihood using an optimization algorithm (gradient descent, conjugate gradient descent etc.)
\item \textbf{Predict:} Provide an estimate of the yield for time step $ t = 1$ using:

\begin{align*}
\hat{\mu}_{0}(\tau*)&=\mathbb{E}(\y_1(\tau*)|\Y_{0})\\
					&=K(\tau*, \tau|\hat{\rho}_{0}). [K(\tau,\tau|\hat{\rho}_{0}) + \hat{\sigma}_{0}^2. \bm{I}]^{-1}.\y_{0}(\tau).
\end{align*}
The predictive interval for time point $1$ can be provided using following distribution
$$
\mu_1|\Y_0 \sim \bf{N}_m(\hat{\mu}_0,\hat{\mathbf{K}}_{0}).
$$
\item \textbf{Update}:
\begin{enumerate}
\item Update the posterior covariance function as:

\begin{equation*}
\resizebox{0.4\textwidth}{!}{
$\hat{\mathbf{K}}_{updated} = K(\tau_*, \tau_*|\hat{\rho}_0) - K(\tau_*, \tau|\hat{\rho}_0). [K(\tau,\tau|\hat{\rho}_0) + \hat{\sigma}_{0}^2. \bm{I}]^{-1}. K(\tau, \tau_*|\hat{\rho}_0).$}
\end{equation*}
\item Update the posterior mean function as:

\resizebox{0.9\linewidth}{!}{
\begin{minipage}{\linewidth}
\begin{align*}
\hat{\mu}_{updated}(\tau*)&=\mathbb{E}(\mu_0(\tau*)|\Y_0)\\
&= K(\tau*, \tau|\hat{\rho}_0). [K(\tau,\tau|\hat{\rho}_0) + \hat{\sigma}_{0}^2. \bm{I}]^{-1}\big(y_1 - \hat{\mu}_{0}\big)
\end{align*}
\end{minipage}
}

which is the mean function associated with the \textbf{Hyper-parameter Estimation} step for time step $t=1$ . 
\end{enumerate} 

\end{enumerate}

\vspace{2mm}

\noindent \textbf{Time step $t\geq 1$}:
\begin{enumerate}
\item \textbf{Hyper-parameter Estimation:} Estimate hyper-parameters $\hat{\theta}_{t}$ of the Gaussian Process $\y_t|\Y_{t-1} \sim \bm{N}_m(\hat{\mu}_{updated}(\tau),\mathbf{K}+\sigma_{t}^2\bm{I}_m)$. Here $\mathbf{K}=K(\tau,\tau*|\rho_t)$ and $\theta_t=(\rho_t,\sigma_{t})$ are the hyper-parameters at time step $t$. The hyper-parameters are obtained by maximizing the marginal log-likelihood using an optimization algorithm.

\item \textbf{Predict:} Provide an estimate of the yield for time step $t + 1$ using:
\begin{align*}
\hat{\mu}_{t}(\tau*) &=\mathbb{E}(\y_{t+1}(\tau*)|\Y_{t})\\
			&=K(\tau*, \tau|\hat{\rho}_{t}). [K(\tau,\tau|\hat{\rho}_{t}) + \hat{\sigma}_{t}^2. \bm{I}]^{-1}.\y_{t}(\tau).
\end{align*}
The predictive interval for time point $t+1$ can be provided using following process
$$
\y_{t+1}|\Y_{t} \sim \bm{N}_m(\hat{\mu}_{t},\hat{\mathbf{K}}_{t}).
$$
\item \textbf{Update:}
\begin{enumerate}
\item Update the posterior covariance function as:
\begin{equation*}
\resizebox{0.4\textwidth}{!}{
$\hat{\mathbf{K}}_{updated} = K(\tau_*, \tau_*|\hat{\rho}_{t}) - K(\tau_*, \tau|\hat{\rho}_{t}). [K(\tau,\tau|\hat{\rho}_{t}) + \hat{\sigma}_{t}^2. \bm{I}]^{-1}. K(\tau, \tau_*|\hat{\rho}_{t})$}
\end{equation*}
\item Update the posterior mean function for term $\tau*$ as:

\resizebox{\linewidth}{!}{
\begin{minipage}{\linewidth}
\begin{align*}
\hat{\mu}_{updated}(\tau*) &=\mathbb{E}(\mu_t(\tau*)|\Y_t)\\
&=\hat{\mu}_t(\tau*)+K(\tau*, \tau|\hat{\rho}_t). [K(\tau,\tau|\hat{\rho}_t) + \hat{\sigma}_{t}^2. \bm{I}]^{-1}.\big(\y_{t + 1} -\hat{\mu}_t(\tau)\big),
\end{align*}
\end{minipage}
}
which is mean function for the \textbf{Hyper-parameter Estimation} step of the subsequent iteration. 
\end{enumerate}
\end{enumerate}

\noindent The covariance function to use with the algorithm is a modeling decision and is problem specific. See \cite{kernel_cookbook} and \cite{Rasmussen2005} for guidelines. For the data used in this study a combination of a linear kernel and a squared exponential (Radial Basis Function) kernel produced good results.

\noindent \textbf{\textit{Remark 1}}: Note that in this dynamic process, the posterior of last time ($t-1$) is being considered as a prior-predictive process for next time point $t$. 

\noindent \textbf{\textit{Remark 2}}: In this algorithm, we are estimating hyper-parameter at every stage. This is feasible because $m$ is small in our case. However, this may not be the possible, in many practical problems.

\subsection{Relationship between Dynamic Gaussian Process and Optimal Bayesian Filter}

\cite{smith1981multiparameter} proposes a Bayesian framework for dynamic models where the prior ($\pi$) at time step $t$, has a power law form:
\begin{equation}\label{eqn:power_law_prior}
\pi(\y_t) \propto p(\y_t|\Y_{t-1}, \delta_t),
\end{equation}
\noindent where:
\begin{itemize}
\item $\y_t$ represents our prior at time step $t$
\item $\delta_t$ represents the power at time step $t$
\end{itemize}
The main idea behind the power filter approach is to propagate information from one time step to the next. The posterior density at stage $t$ is given by:
\begin{equation}
\pi(\y_t) \propto f(\y_t|\Y_{t-1}) . \big[\pi_{t|t-1}(\y_t)\big]^{\delta_t} \quad such\ that\ 0\leq\delta_t\leq 1,
\end{equation} 
\noindent where:
\begin{itemize}
\item $f(\y_t|\Y_{t -1})$ represents the likelihood
\item $\big[\pi_{t|t-1}(\y_t)\big]^{\delta_t}$ represents the prior
\end{itemize}
When $\delta_t = 1$ and the likelihood and the prior are Multivariate Gaussian, then we obtain the Dynamic GP. \cite{das2013dynamic} develop the power filter for dynamic generalized linear models. They show that the power filter model yields an efficient information processing rule in a dynamic model setting.

\section{Validation Methodology} \label{sec:test_methodology}
The data for this study came from the website of the US Department of Treasury (\cite{treas_data}). The data represents over 10 years of yield curve data ( February 2006 through February 2017). A rolling window was used to train and test the performance of the methods on the proposed dataset. The details of this procedure are as follows. Starting with the yield curve data for the first day, we select a batch of data to be used for training the method used for yield curve forecasting. We then use the developed model to score the first data point after the batch of data points used as training data. For training the next batch, we remove the first data point and include the first test point in the training set. As we repeat this process, we move through the dataset, forecasting one test point at a time. Forecasting using the multivariate time series method for either the Nelson-Siegel parameters or the term yield forecasts themselves are not Bayesian methods. We used 250 days of data for training for these methods. This corresponds to about a year of data. This implies one year of data is used to train the time series methods to forecast a test point. Gaussian Process regression is a Bayesian method. Section \ref{sec:dgp} provides the details of training and forecasting using the Dynamic Gaussian Process Model . 
\section{Results and Discussion}\label{sec:results}
This study examines data over a ten year period. The Root Mean Square Error was used as the metric to assess the performance of the method. The Root Mean Square Error is defined by:
\begin{equation}
\textit{RMSE} = \sqrt{\frac{\sum_{i=1}^{i=N}\sum_{\tau = 1}^{\tau = 11} (\hat{\y}[\tau, i] -y[\tau, i])^2}{N}},
\end{equation}
\noindent where:
\begin{itemize}
\item $\hat{\y}[\tau, i]$ is the estimated yield for day $i$ associated with term $\tau$
\item $\y[\tau, i]$ is the actual yield for day $i$ associated with term $\tau$
\item $N$ is the number of yield curves that are estimated using the procedure
\item There are 11 terms in each yield curve. 
\end{itemize}
A summarized view of the results for this ten year period are shown in Table \ref{tab:term_rmse}.
\begin{table}[ht]
\centering
\begin{tabular}{|l|c|c|c|}
  \hline
 Term & GP & MVTS & TSNS \\ 
  \hline
1 Month & 0.104 & 0.088 & 0.121 \\ 
  3 Months & 0.071 & 0.066 & 0.080 \\ 
  6 Months & 0.054 & 0.047 & 0.088 \\ 
  1 Year & 0.047 & 0.043 & 0.085 \\ 
  2 Years & 0.052 & 0.055 & 0.088 \\ 
  3 Years & 0.058 & 0.061 & 0.114 \\ 
  5 years & 0.065 & 0.068 & 0.126 \\ 
  7 Years & 0.065 & 0.070 & 0.149 \\ 
  10 Years & 0.063 & 0.067 & 0.197 \\ 
  20 Years & 0.061 & 0.065 & 0.977 \\ 
  30 Years & 0.060 & 0.063 & 10.838 \\ 
   \hline
\end{tabular}
\caption{RMSE for term structures for all methods}
\label{tab:term_rmse}
\end{table}
\noindent An inspection of Table \ref{tab:term_rmse} shows that the Nelson Siegel model based time series does relatively poorly in comparison to the multivariate time series method and the dynamic GP.  We examine the performance of these methods over three time durations - short term structures, medium term structures and long term structures. The short term structure included term structures upto 1 year. The medium term structures consists of bonds with maturities of 2 years, 3 years and 5 years. The long term structure category consists of bonds that mature at 7 years, 10 years, 20 years and 30 years. The multivariate time series appears to do well in the short term region of the yield curve while the dynamic GP method does well in the medium and long term regions of the yield curve. 

\begin{figure}
     \begin{center}
        \subfigure[Estimates for Feb 06, 2008]{%
            \label{fig:d500sample}
            \includegraphics[width=0.4\textwidth]{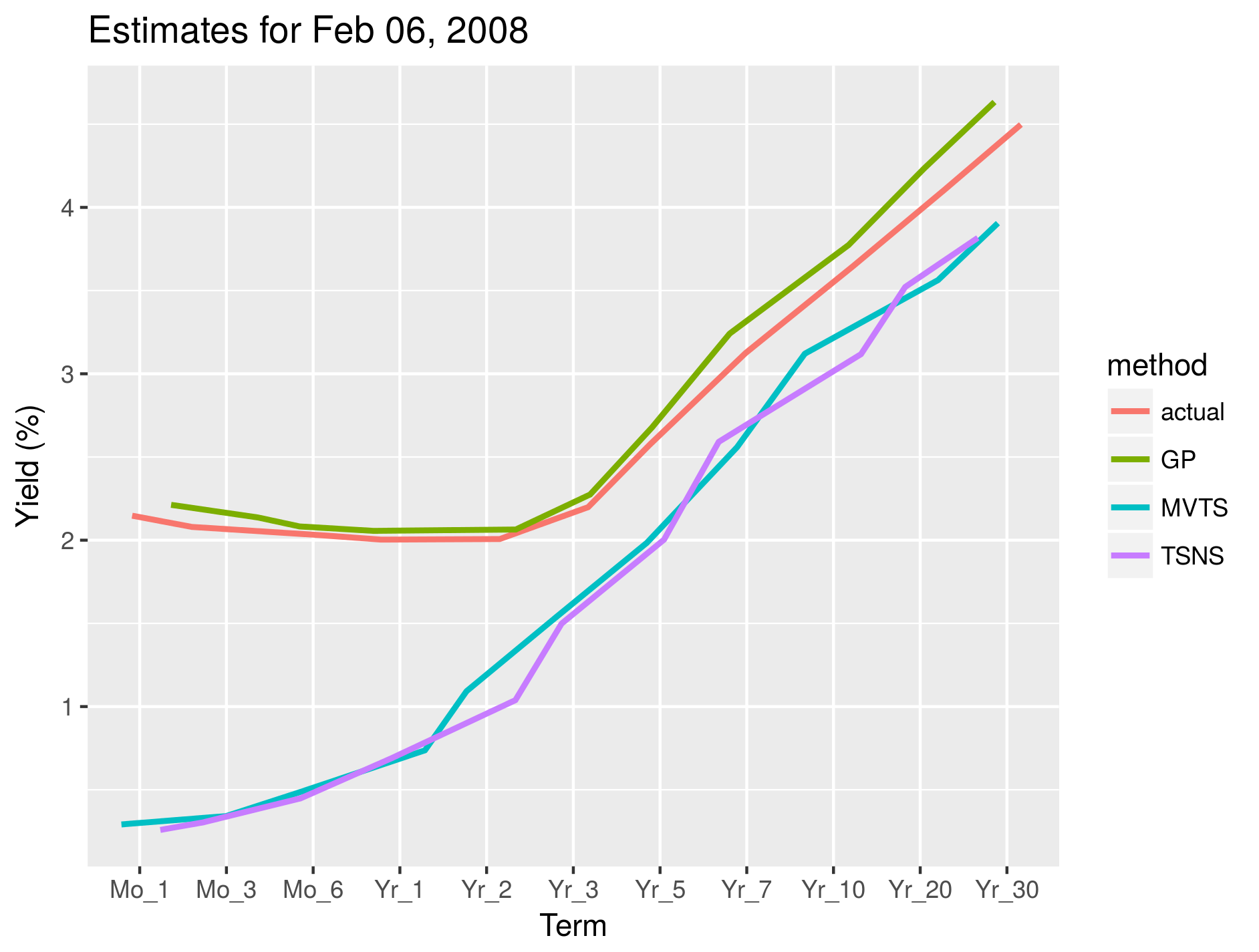}
        }%
        \subfigure[Estimates for Feb 10, 2010]{%
           \label{fig:d1000sample}
           \includegraphics[width=0.4\textwidth]{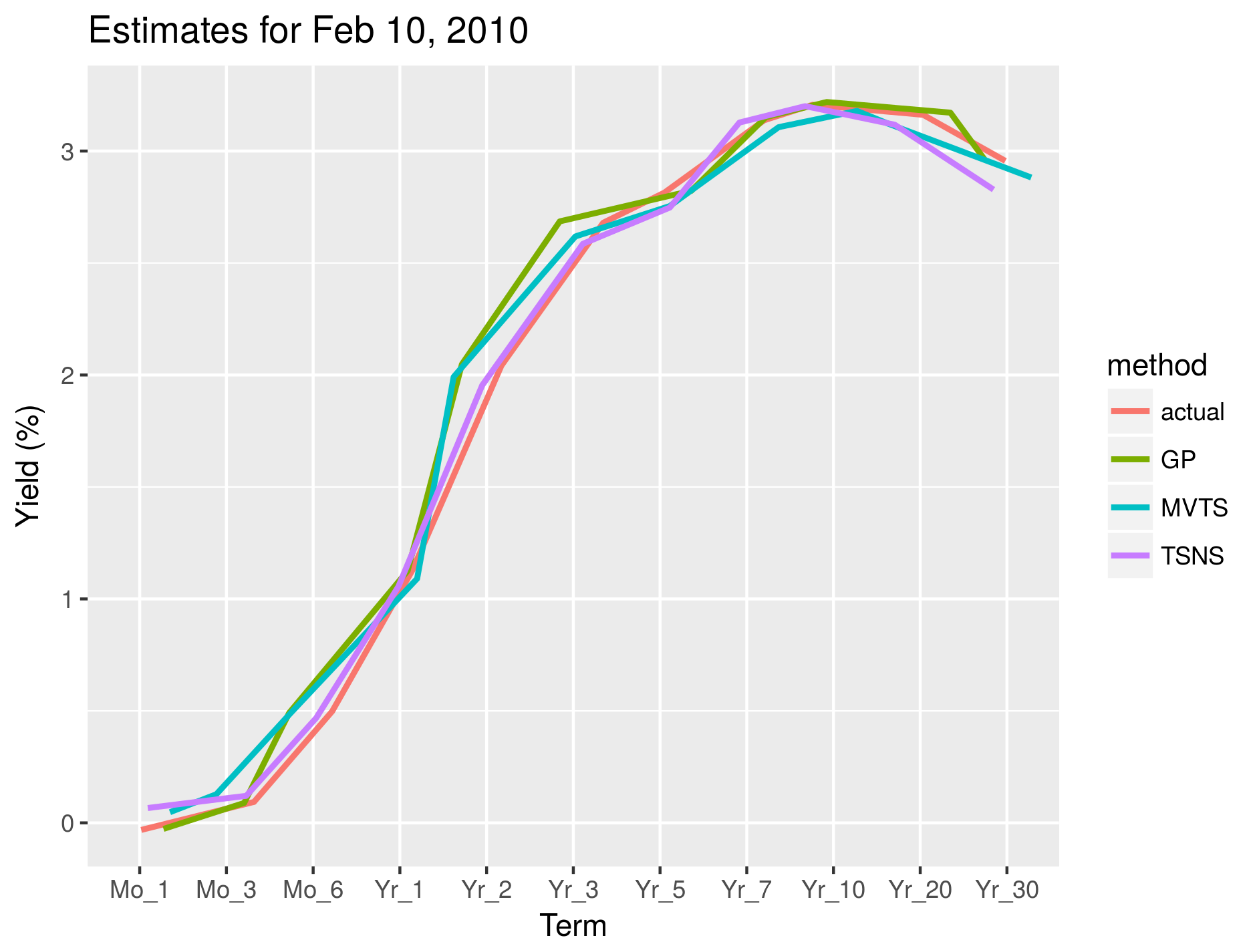}
        }\\ 
        \subfigure[Estimates for Feb 08, 2012]{%
            \label{fig:d188sample}
            \includegraphics[width=0.4\textwidth]{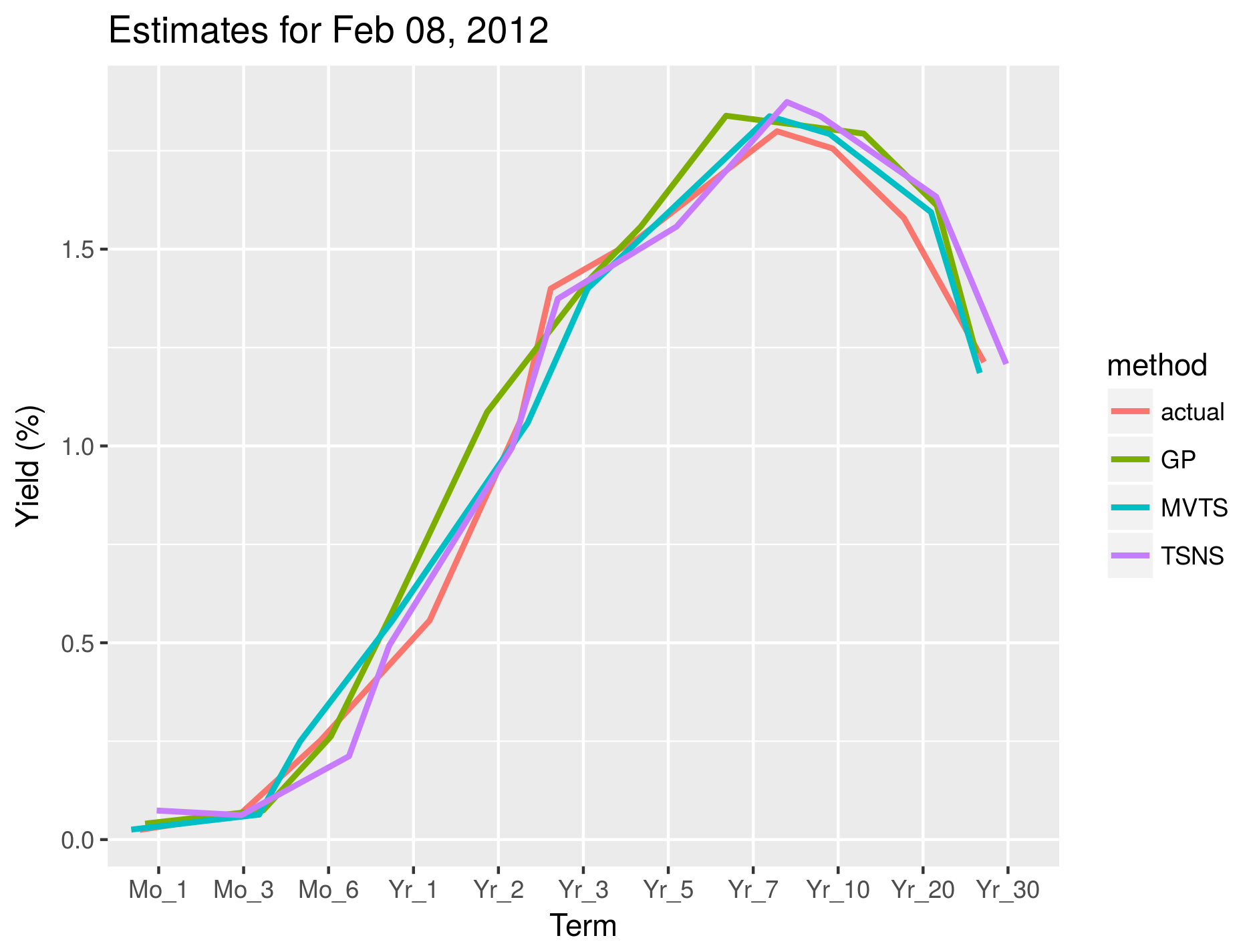}
        }%
        \subfigure[Estimates for Feb 07, 2014]{%
            \label{fig:d2000sample}
            \includegraphics[width=0.4\textwidth]{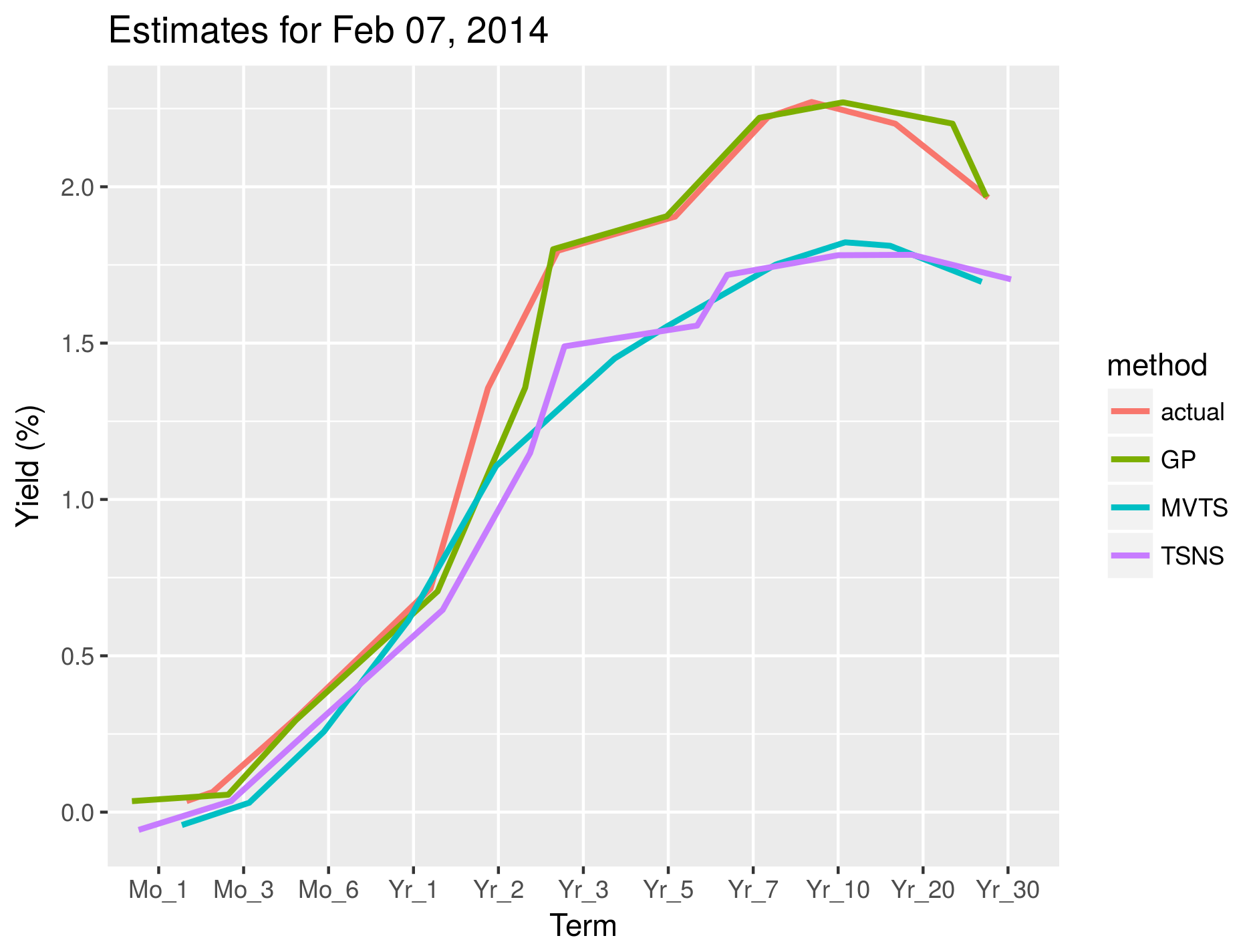}
        }%
    \end{center}
    \caption{%
Estimates for a Sample of the Data Using the Methods Discussed
     }%
   \label{fig:sample_of_estimates}
\end{figure}

\begin{figure}[H]
     \begin{center}
        \subfigure[1 Month Performance]{%
            \label{fig:1moperf}
            \includegraphics[width=0.4\textwidth]{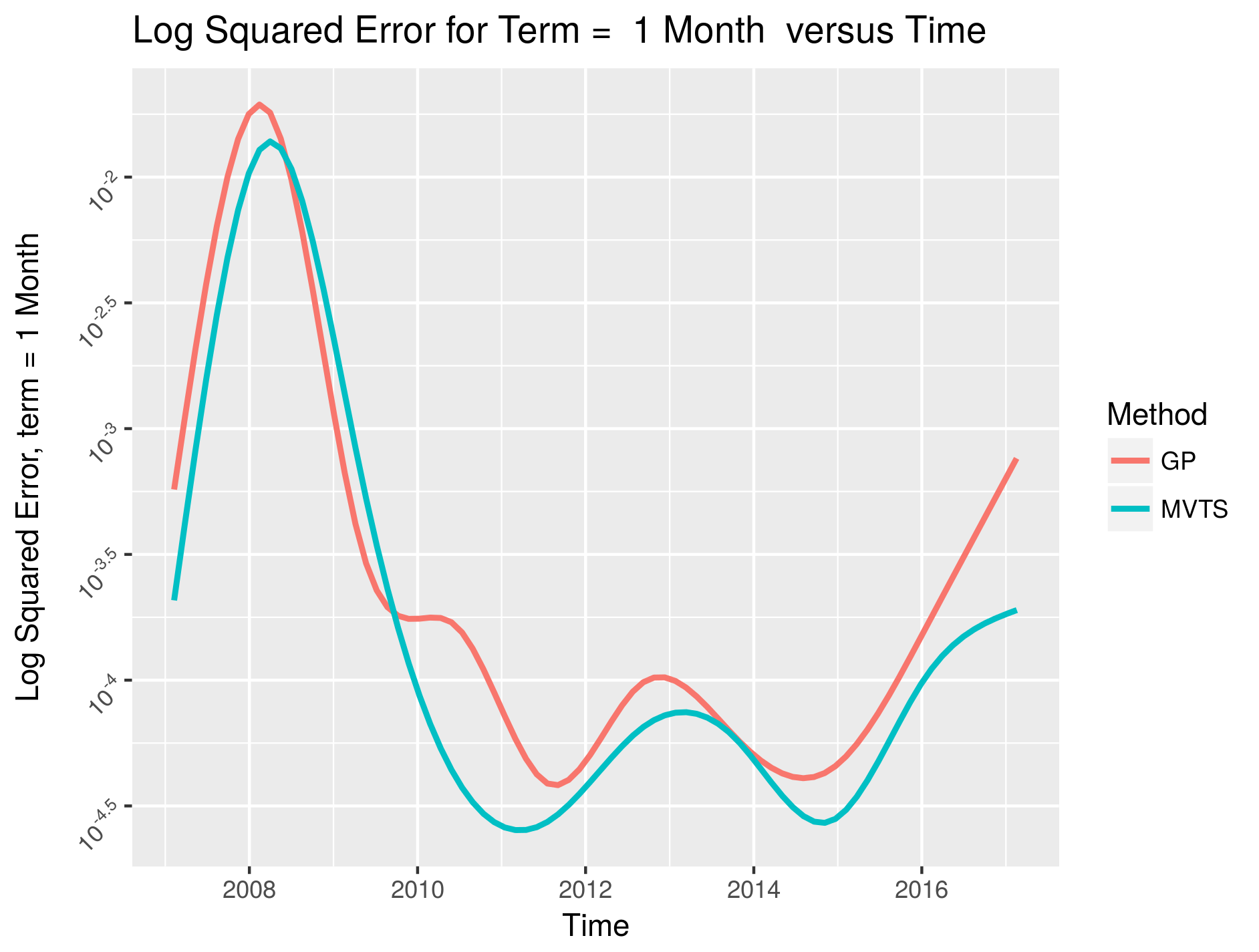}
        }%
        \subfigure[3 Month Performance]{%
           \label{fig:3moperf}
           \includegraphics[width=0.4\textwidth]{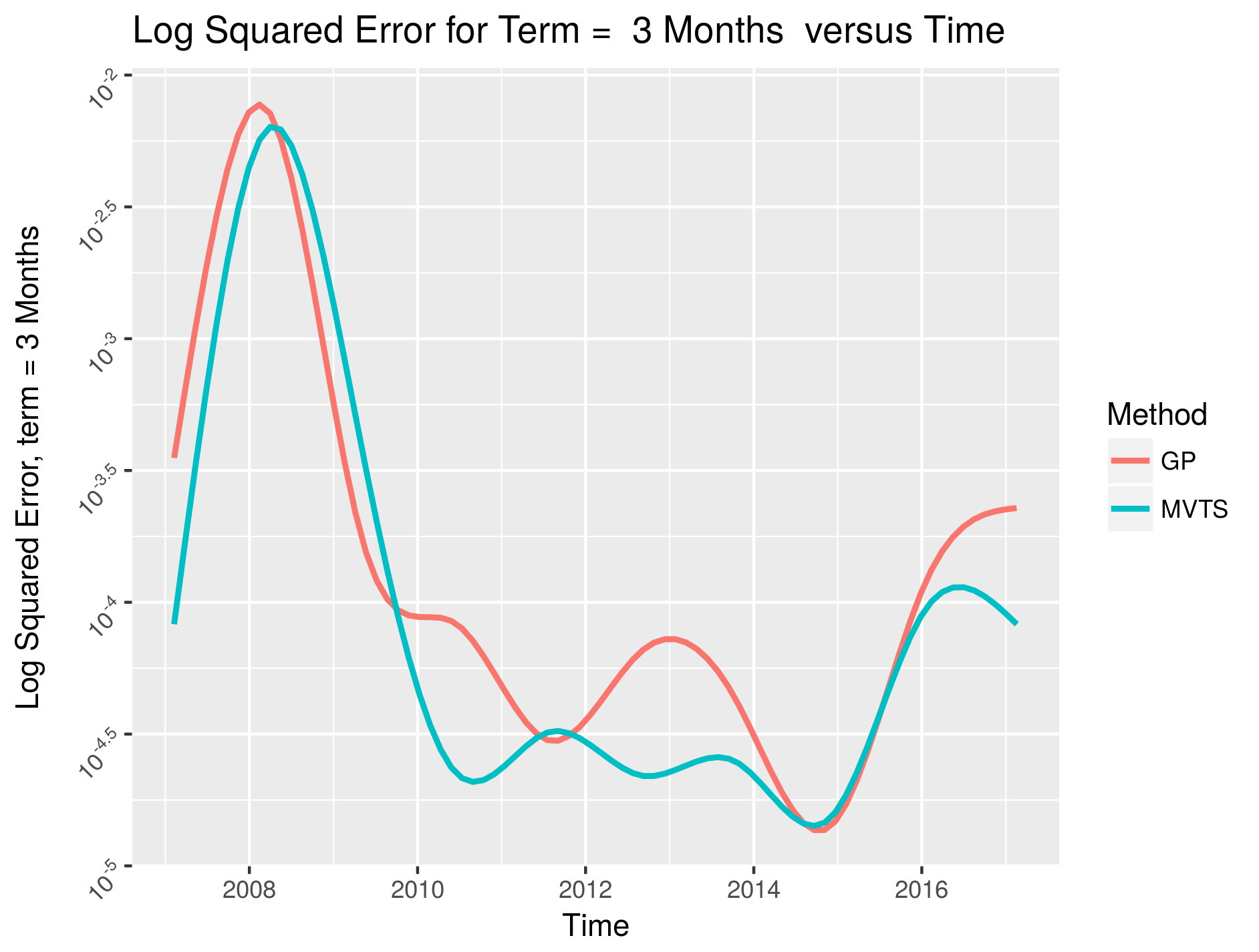}
        }\\ 
        \subfigure[6 Month Performance]{%
            \label{fig:6moperf}
            \includegraphics[width=0.4\textwidth]{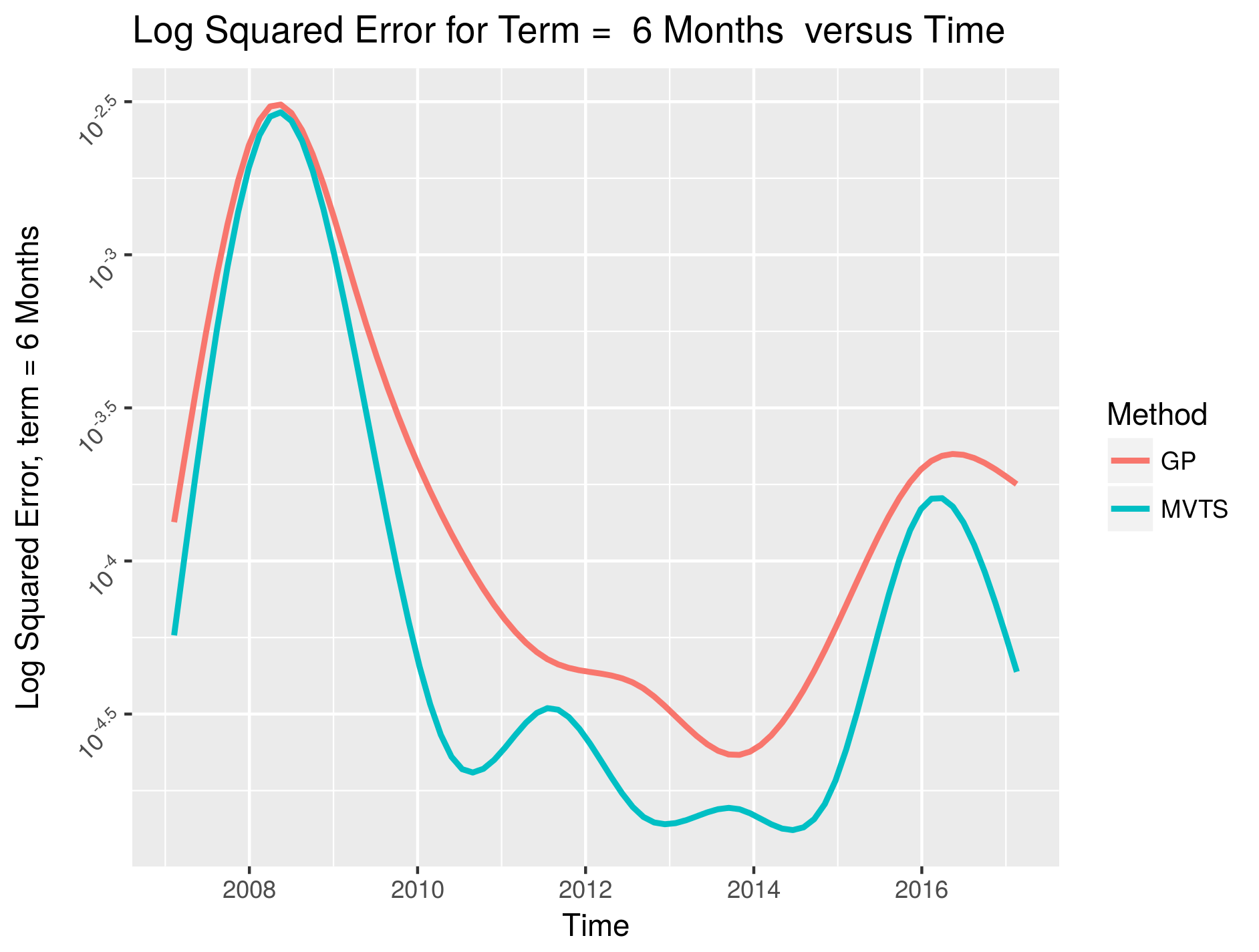}
        }%
        \subfigure[1 Year Performance]{%
            \label{fig:1yrperf}
            \includegraphics[width=0.4\textwidth]{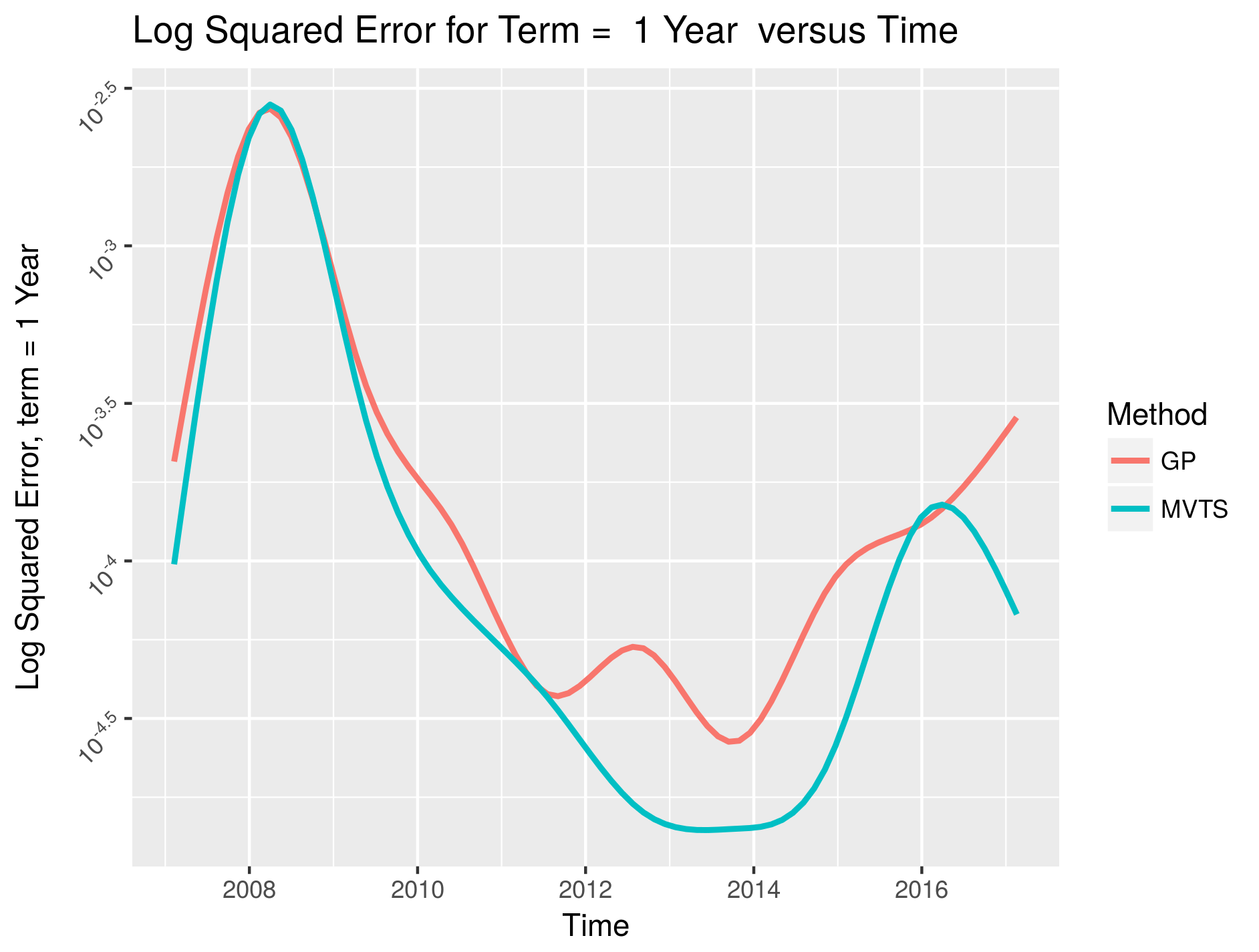}
        }%
    \end{center}
    \caption{%
	Short Term Performance - GP versus MVTS over a 10 Year Period
     }%
   \label{fig:short_term_estimates}
\end{figure}

\begin{figure}[H]
     \begin{center}
        \subfigure[2 Year Performance]{%
            \label{fig:2yrperf}
            \includegraphics[width=0.4\textwidth]{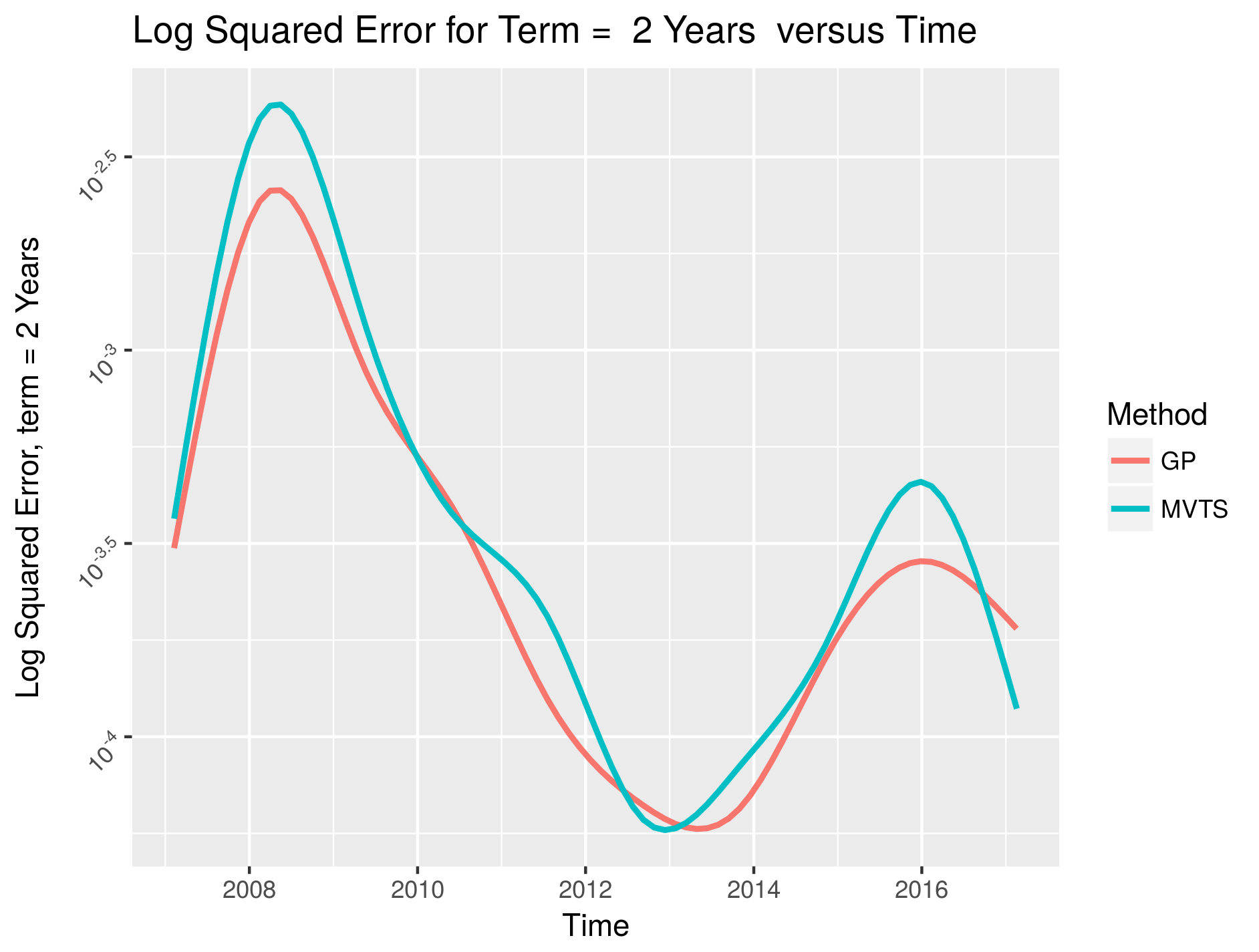}
        }%
        \subfigure[3 Year Performance]{%
           \label{fig:3yrperf}
           \includegraphics[width=0.4\textwidth]{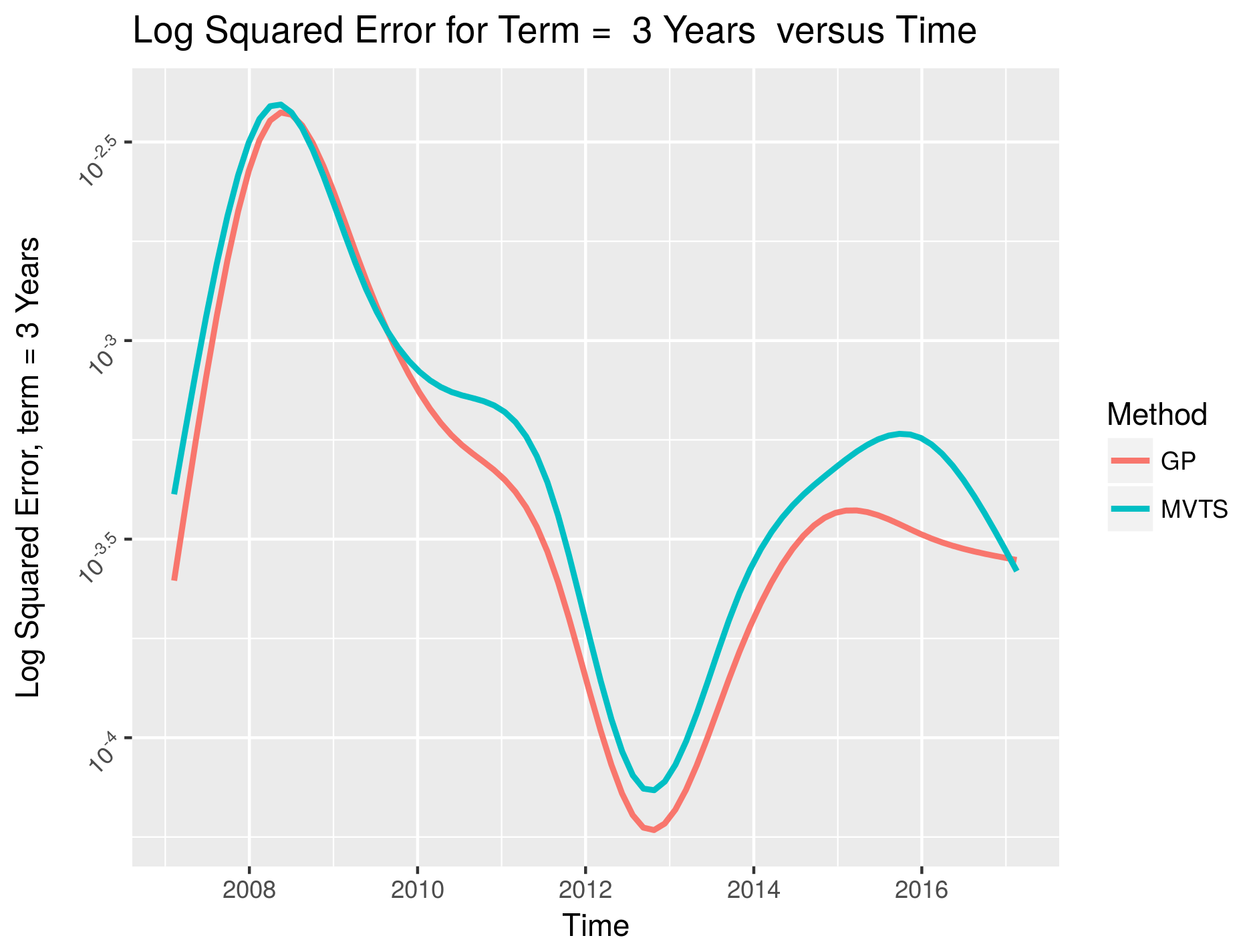}
        }\\ 
        \subfigure[5 Year Performance]{%
            \label{fig:5yrperf}
            \includegraphics[width=0.4\textwidth]{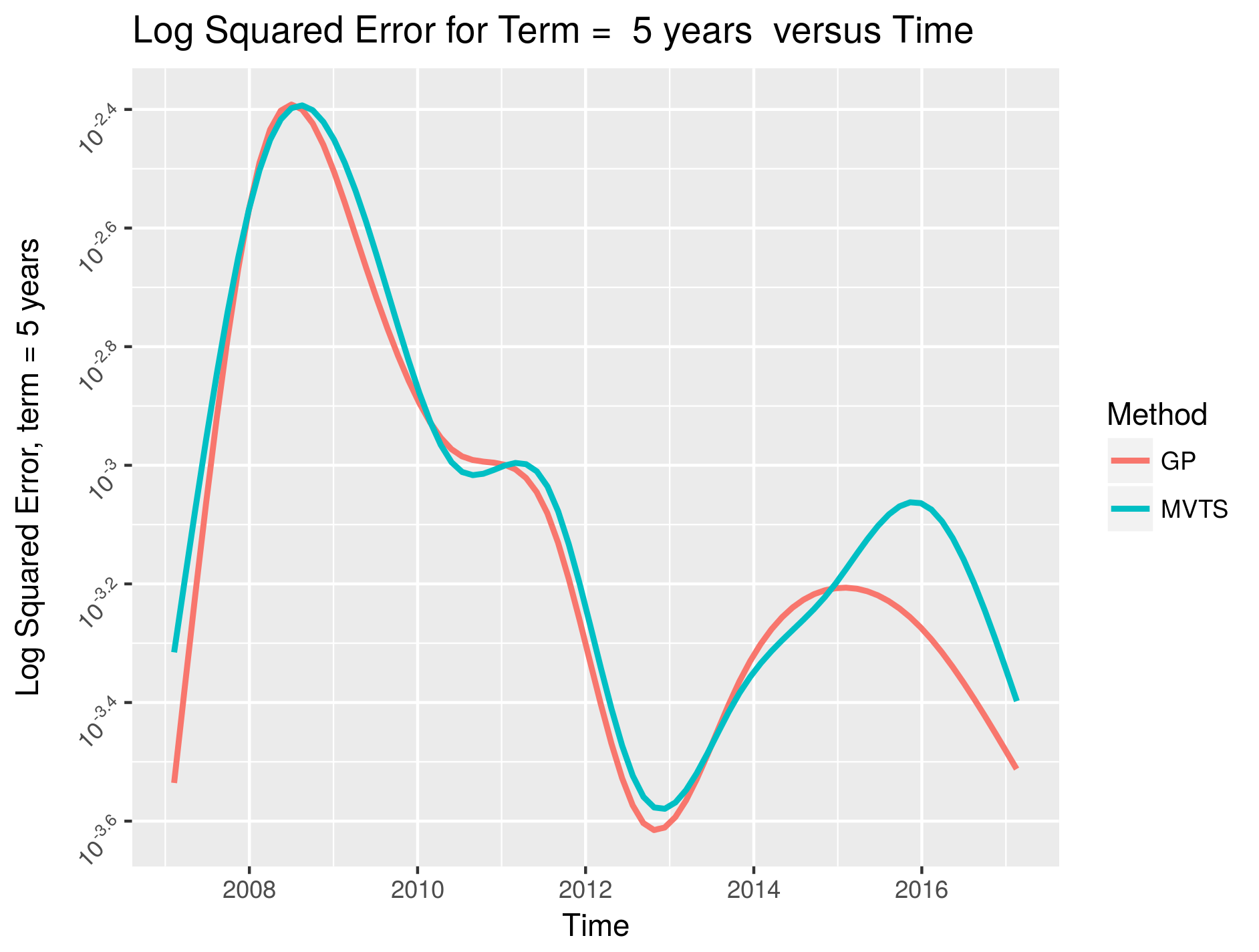}
        }%

    \end{center}
    \caption{%
	Medium Term Performance - GP versus MVTS over a 10 Year Period
     }%
   \label{fig:medium_term_estimates}
\end{figure}

\begin{figure}[H]
     \begin{center}
        \subfigure[7 Year Performance]{%
            \label{fig:7yrperf}
            \includegraphics[width=0.4\textwidth]{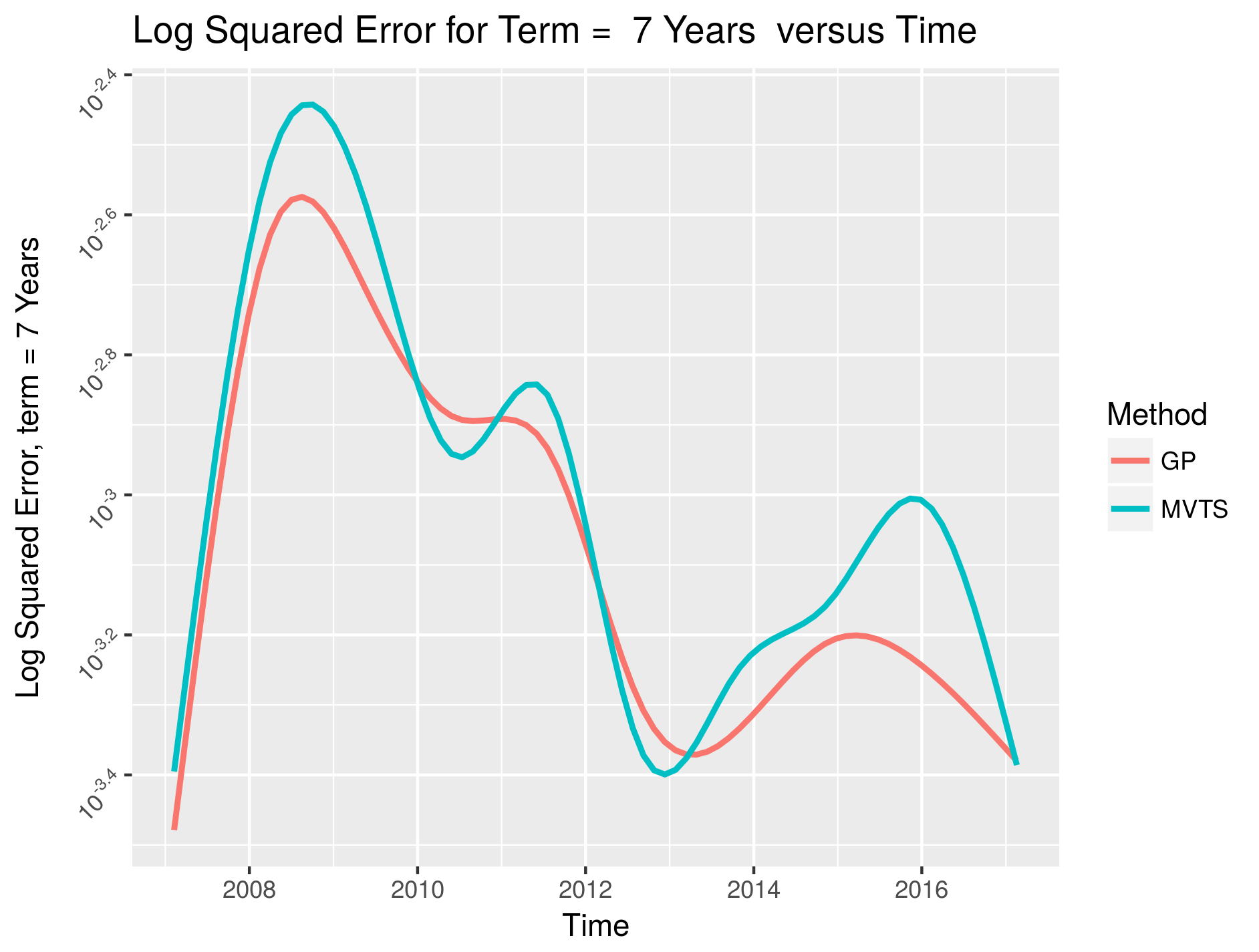}
        }%
        \subfigure[10 Year Performance]{%
           \label{fig:10yrperf}
           \includegraphics[width=0.4\textwidth]{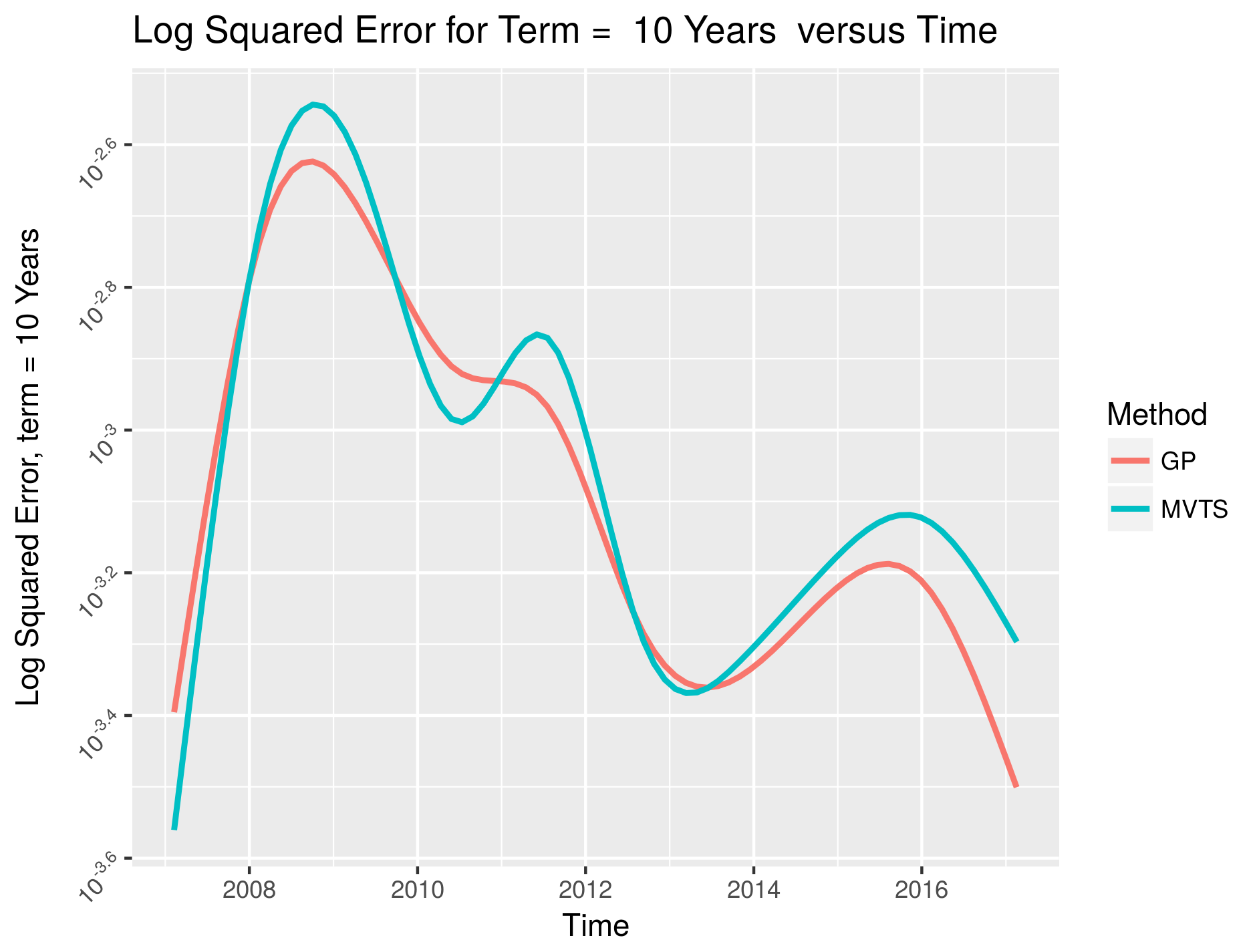}
        }\\ 
        \subfigure[20 Year Performance]{%
            \label{fig:20yrperf}
            \includegraphics[width=0.4\textwidth]{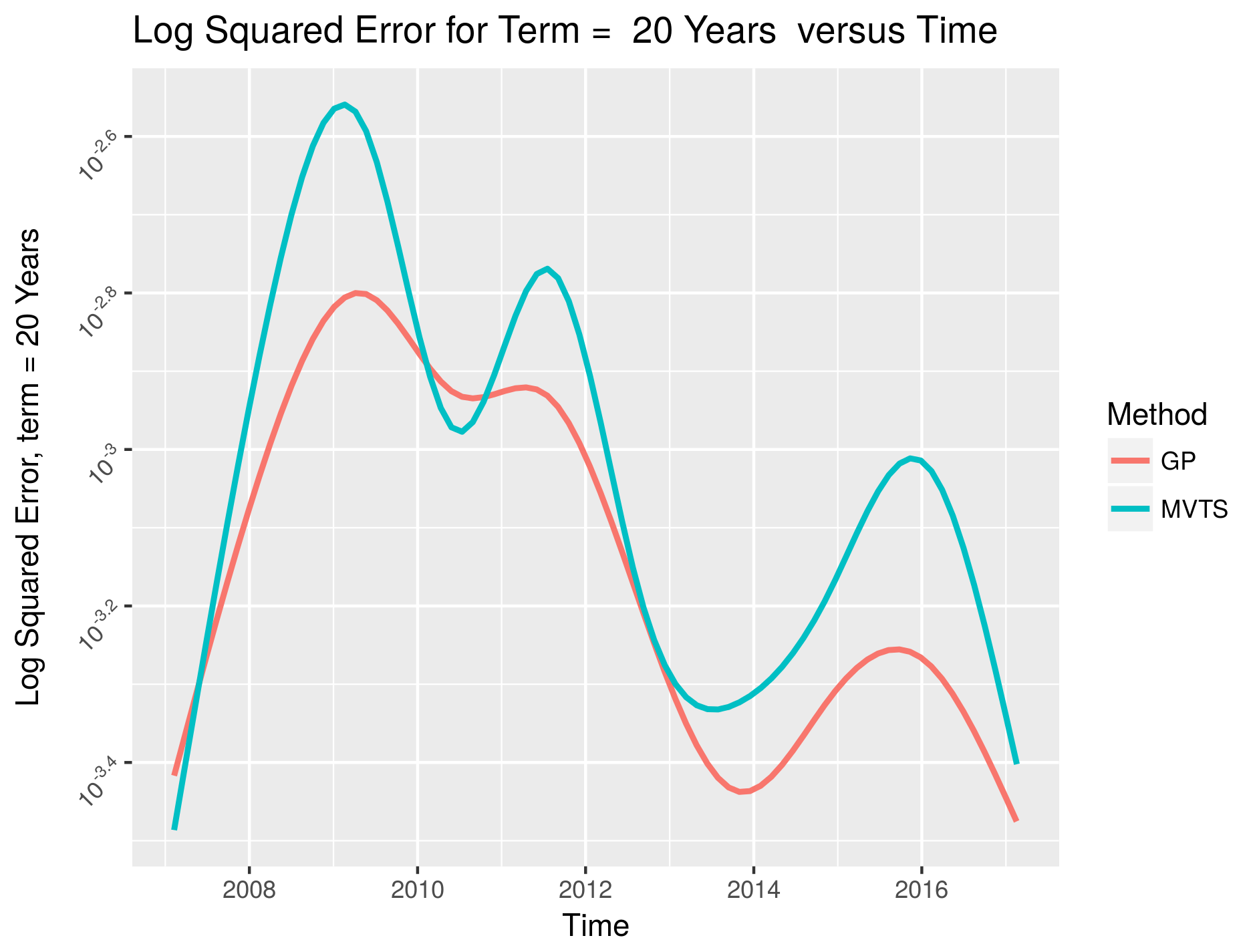}
        }%
        \subfigure[30 Year Performance]{%
            \label{fig:30yrperf}
            \includegraphics[width=0.4\textwidth]{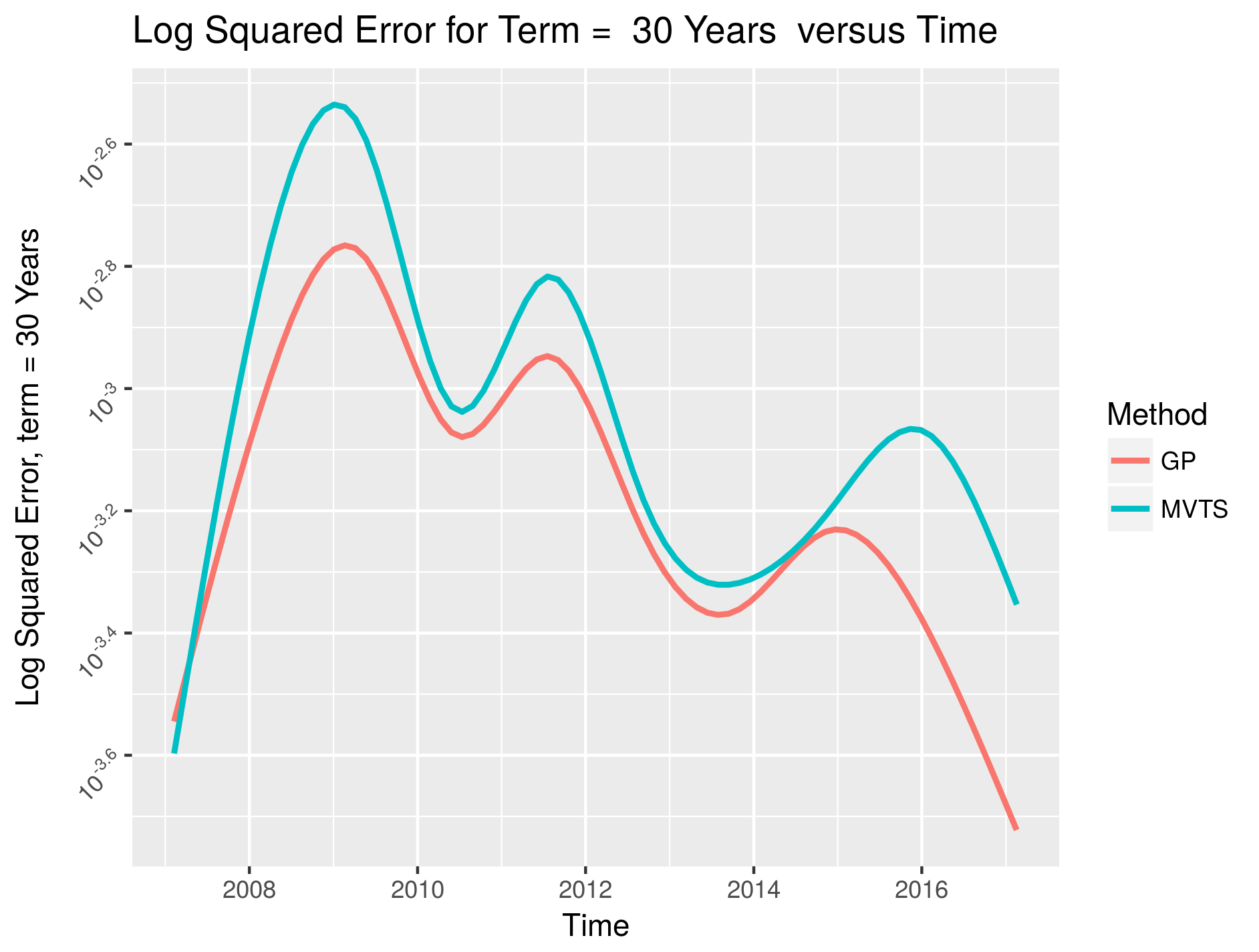}
        }%
    \end{center}
    \caption{%
	Long Term Performance - GP versus MVTS over a 10 Year Period
     }%
   \label{fig:long_term_estimates}
\end{figure}

Figure \ref{fig:sample_of_estimates} shows estimates using the techniques discussed for a sample of days in the dataset. The data for days $500$, $1000$, $1500$, and $2000$ were used for this illustration. This provides a cross-sectional view of the estimates from the various methods used in this study. Note that Figure \ref{fig:sample_of_estimates} also provides the actual yield associated with the term-structures on these days. On some days like February 10, 2010 and February 08, 2012, the estimates from all methods are close to the actual. On other days the estimates may be quite different. However an inspection of figure \ref{fig:sample_of_estimates} shows that the estimates from the dynamic GP agree quite well with the actual yields over the 10 year period considered for this study. Estimates from the methods are usually quite close, so a small amount of Gaussian noise (jitter) was added to discriminate the curves in Figure \ref{fig:sample_of_estimates}. As evident from Table \ref{tab:term_rmse}, the performance of the Nelson-Siegel based method is inferior to that of the multivariate time series method and the dynamic GP method. Therefore, in the analysis that follows, we limit our discussion to the comparison of the dynamic GP method and the multivariate time series method. Figure \ref{fig:short_term_estimates} through Figure \ref{fig:long_term_estimates} show the squared error of the estimates associated with the dynamic GP method and the multivariate time series methods.

\noindent The short term performance of the dynamic GP and the multivariate time series methods over the 10 year period  is shown in Figure \ref{fig:short_term_estimates}. A review of Figure \ref{fig:short_term_estimates} shows that the performance of the the multivariate time series method is in general better than the dynamic GP method in the short term region of the yield curve.

\noindent The medium term performance of the dynamic GP and the multivariate time series methods over a 10 year period is shown in Figure \ref{fig:medium_term_estimates}. The long term performance of these methods is shown in \ref{fig:long_term_estimates}. An analysis of the medium term and long term performance curves shows that the dynamic GP performs better than the multivariate time series method in the medium and short term structure regions. This is consistent with summarized RMSE over the 10 year period in Table \ref{tab:term_rmse}.

The \texttt{GPy python} package \cite{gpy2014} was used for developing the Gaussian Process models reported in 
this work. The \texttt{vars R} package\cite{rpkg_var} was used to model the time series based methods to forecast the Nelson-Siegel coefficients or the yield curve term rates.

\section{Conclusion}\label{sec:conclusion}
Gaussian processes have been used for functional data analysis in several domains (see \cite{Rasmussen2005}). The results of this study suggest that they can be used for yield curve forecasting. The nature of yield curve data is such that there is more data in the short and medium term structure regions than the long term structure regions . This makes long term forecasts challenging. This study revealed that the proposed dynamic GP method can forecast this region of the yield curve well. Analysts could use a mix of methods to forecast the yield curve. The data for this study spans a large time interval - over ten years. The results of this study indicate that the multivariate time series approach is more accurate for forecasting the short term structures, while the proposed dynamic Gaussian Process based method is a better choice for the medium and long term structures associated with the yield curve. The proposed method has been applied to a forecasting problem in the financial domain, however, this method can be applied to other domains as well. Demand forecasting is a common business requirement. In an IT data center, we might interested in forecasting the hourly number of user requests serviced by a group of computers. The hourly energy demand might be of interest to an electrical utility company. In summary, we believe that the dynamic Gaussian Process model could be useful in other application domains too.

\medskip
\nocite{*}
\bibliographystyle{plainnat}

\bibliography{sample}

\begin{thebibliography}{25}
\providecommand{\natexlab}[1]{#1}
\providecommand{\url}[1]{\texttt{#1}}
\expandafter\ifx\csname urlstyle\endcsname\relax
  \providecommand{\doi}[1]{doi: #1}\else
  \providecommand{\doi}{doi: \begingroup \urlstyle{rm}\Url}\fi

\bibitem[Anderson(1984)]{Anderson1984}
T.~W. Anderson.
\newblock \emph{An Introduction to Multivariate Statistical Analysis}.
\newblock Wiley, 1984.

\bibitem[Chen et~al.(2000)Chen, Ibrahim, and Shao]{chen2000power}
Ming-Hui Chen, Joseph~G Ibrahim, and Qi-Man Shao.
\newblock Power prior distributions for generalized linear models.
\newblock \emph{Journal of Statistical Planning and Inference}, 84\penalty0
  (1):\penalty0 121--137, 2000.

\bibitem[Chen and Niu(2014)]{Chen_Niu_2014}
Ying Chen and Linlin Niu.
\newblock Adaptive dynamic nelson-siegel term structure model with
  applications.
\newblock \emph{Journal of Econometrics}, 180\penalty0 (1):\penalty0 98--115,
  2014.

\bibitem[Das et~al.(2016)Das, Lahiri, and Das]{das2016understanding}
Purba Das, Ananya Lahiri, and Sourish Das.
\newblock Understanding sea ice melting via functional data analysis.
\newblock \emph{arXiv preprint arXiv:1610.07024}, 2016.

\bibitem[Das and Dey(2013)]{das2013dynamic}
Sourish Das and Dipak~K Dey.
\newblock On dynamic generalized linear models with applications.
\newblock \emph{Methodology and Computing in Applied Probability}, pages 1--15,
  2013.

\bibitem[Diebold and Li(2006)]{Diebold_Li_2006}
Francis Diebold and Canlin Li.
\newblock Forecasting the term structure of government bond yields.
\newblock \emph{Journal of Econometrics}, 130\penalty0 (1):\penalty0 337--364,
  2006.

\bibitem[Diebold and Rudebusch(2013)]{diebold2013yield}
Francis~X Diebold and Glenn~D Rudebusch.
\newblock \emph{Yield Curve Modeling and Forecasting: The Dynamic Nelson-Siegel
  Approach}.
\newblock Princeton University Press, 2013.

\bibitem[Duvenaud(2017)]{kernel_cookbook}
David Duvenaud.
\newblock \emph{{Kernel Cookbook} Kernel Cookbook}, 2017.
\newblock URL \url{http://www.cs.toronto.edu/~duvenaud/cookbook/index.html}.

\bibitem[Friedman et~al.(2001)Friedman, Hastie, and
  Tibshirani]{friedman2001elements}
Jerome Friedman, Trevor Hastie, and Robert Tibshirani.
\newblock \emph{The elements of statistical learning}, volume~1.
\newblock Springer series in statistics Springer, Berlin, 2001.

\bibitem[{GPy}(2012--2014)]{gpy2014}
{GPy}.
\newblock {GPy}: A gaussian process framework in python.
\newblock \url{http://github.com/SheffieldML/GPy}, 2012--2014.

\bibitem[Gupta and Ibrahim(2009)]{Gupta_Ibrahim_2009}
Mayetri Gupta and Joseph~G. Ibrahim.
\newblock An information matrix prior for bayesian analysis in generalized
  linear models with high dimensional data.
\newblock \emph{Stat Sinica}, 2009.

\bibitem[Ibrahim et~al.(2003)Ibrahim, Chen, and Sinha]{ibrahim2003optimality}
Joseph~G Ibrahim, Ming-Hui Chen, and Debajyoti Sinha.
\newblock On optimality properties of the power prior.
\newblock \emph{Journal of the American Statistical Association}, 98\penalty0
  (461):\penalty0 204--213, 2003.

\bibitem[Meinhold and Singpurwalla(1983)]{MeinholdSingpurwala1983}
Richard~J. Meinhold and Nozer~D. Singpurwalla.
\newblock Understanding the kalman filter.
\newblock \emph{The American Statistician}, 37\penalty0 (2):\penalty0 123--127,
  1983.

\bibitem[Nelson and Siegel(1987)]{Nelson_Siegel_1987}
Charles~R. Nelson and Andrew~F. Siegel.
\newblock Parsimonious modeling of yield curve.
\newblock \emph{The Journal of Business}, 60\penalty0 (4):\penalty0 473--489,
  1987.

\bibitem[Nielsen(2017)]{yield_curve_imp}
Barry Nielsen.
\newblock \emph{{Bond Yield Curve Holds Predictive Powers} Treasury Rates},
  2017.
\newblock URL
  \url{http://www.investopedia.com/articles/economics/08/yield-curve.asp}.

\bibitem[Pfaff(2008{\natexlab{a}})]{book_var}
B.~Pfaff.
\newblock \emph{Analysis of Integrated and Cointegrated Time Series with R}.
\newblock Springer, New York, second edition, 2008{\natexlab{a}}.
\newblock URL \url{http://www.pfaffikus.de}.
\newblock ISBN 0-387-27960-1.

\bibitem[Pfaff(2008{\natexlab{b}})]{rpkg_var}
Bernhard Pfaff.
\newblock Var, svar and svec models: Implementation within {R} package {vars}.
\newblock \emph{Journal of Statistical Software}, 27\penalty0 (4),
  2008{\natexlab{b}}.
\newblock URL \url{http://www.jstatsoft.org/v27/i04/}.

\bibitem[{R Core Team}(2016)]{}
{R Core Team}.
\newblock \emph{R: A Language and Environment for Statistical Computing}.
\newblock R Foundation for Statistical Computing, Vienna, Austria, 2016.
\newblock URL \url{https://www.R-project.org/}.

\bibitem[Ramsay and Silverman(2002)]{ramsay2002applied}
James~O Ramsay and Bernard~W Silverman.
\newblock \emph{Applied functional data analysis: methods and case studies},
  volume~77.
\newblock Citeseer, 2002.

\bibitem[Rasmussen and Williams(2005)]{Rasmussen2005}
Carl~Edward Rasmussen and Christopher K.~I. Williams.
\newblock \emph{Gaussian Processes for Machine Learning (Adaptive Computation
  and Machine Learning)}.
\newblock The MIT Press, 2005.
\newblock ISBN 026218253X.

\bibitem[Rossum(1995)]{Rossum:1995:PRM:869369}
Guido Rossum.
\newblock Python reference manual.
\newblock Technical report, Amsterdam, The Netherlands, The Netherlands, 1995.

\bibitem[Smith(1981)]{smith1981multiparameter}
JQ~Smith.
\newblock The multiparameter steady model.
\newblock \emph{Journal of the Royal Statistical Society. Series B
  (Methodological)}, pages 256--260, 1981.

\bibitem[Spencer~Hays and Huang(2012)]{Hays_Shen_Huang_2012}
Haipeng~Shen Spencer~Hays and Jianhua~Z. Huang.
\newblock Functional dynamic factor models with applications to yield curve
  forecasting.
\newblock \emph{Annals of Applied Statistics}, 6\penalty0 (3):\penalty0
  870--894, 2012.

\bibitem[West(1986)]{west1986bayesian}
Mike West.
\newblock Bayesian model monitoring.
\newblock \emph{Journal of the Royal Statistical Society. Series B
  (Methodological)}, pages 70--78, 1986.

\bibitem[www.treasury.gov(2017)]{treas_data}
www.treasury.gov.
\newblock \emph{{Treasury Rates} Treasury Rates}, 2017.
\newblock URL
  \url{https://www.treasury.gov/resource-center/data-chart-center/interest-rates/Pages/TextView.aspx?data=yield}.

\end{thebibliography}

\end{document}